\documentclass[runningheads]{llncs}

\usepackage{eccv}

\usepackage{eccvabbrv}

\usepackage{graphicx}
\usepackage{pdfpages}
\usepackage{booktabs}
\usepackage{wrapfig}
\usepackage{kotex}
\usepackage{amsmath} 
\usepackage{amssymb}  
\usepackage{booktabs}
\usepackage{multirow}
\usepackage{kotex}
\usepackage{lipsum}
\usepackage{float}
\usepackage{amsmath}
\usepackage{colortbl}
\usepackage{threeparttable}
\usepackage{graphicx}    
\usepackage{caption} 
\usepackage{booktabs}

\usepackage{xcolor}
\definecolor{mybrown}{rgb}{0.65, 0.16, 0.16} 
\usepackage[accsupp]{axessibility}  

\newcommand{\ourmodel}{Graph-GSReg}

\usepackage{hyperref}

\usepackage{orcidlink}

\begin{document}

\title{Graph-GSReg: Leveraging 3D Scene Graphs for Gaussian Splatting Registration}

\titlerunning{Graph-GSReg}

\author{Jaewon Lee\inst{1}\orcidlink{0009-0003-0507-9870} \and
Mangyu Kong\inst{1}\orcidlink{0009-0006-4281-8814} \and
Euntai Kim\inst{1,2}\thanks{Corresponding author}\orcidlink{0000-0002-0975-8390}}

\authorrunning{J.~Lee et al.}

\institute{Yonsei University, Seoul, Republic of Korea \and Korea Institute of Science and Technology, Seoul, Republic of Korea
\\
\email{\{leejaewon,mangyu0929,etkim\}@yonsei.ac.kr}
\\
\url{https://lee-jaewon.github.io/Graph-GSReg/}
}

\maketitle

\begin{abstract}

Merging multiple 3D Gaussian Splatting (3DGS) scenes into a single unified Gaussian representation is essential for large-scale 3D mapping and long-term map management. Despite its importance, this area remains underexplored, and existing solutions exhibit several limitations.
Learning-based methods attempt direct correspondence between Gaussian primitives and require training on large 3DGS datasets. Image-based optimization methods depend heavily on coarse initialization from generic foundation models and often incur expensive refinement.
We present \ourmodel. Our method constructs a 3D scene graph from a 3DGS and its rendered images, \textit{reformulating 3DGS registration as a graph registration problem}. The proposed 3D scene graph represents each 3DGS at a higher-level representation, enabling a globally consistent understanding of semantic information and structural context for accurate registration.
To further construct a seamless unified scene, we introduce a Self-Supervised Test-Time Optimization. Naively merging two 3D Gaussian scenes often suffers from occlusion artifacts such as hollows and floaters. To alleviate this issue, we refine the merged Gaussians to preserve visual consistency between the original scenes and the merged scene.
We evaluate our method on real and synthetic benchmarks, demonstrating competitive registration accuracy and merged scene rendering quality.

\end{abstract}

\begin{figure}[ht]
  \centering
  \includegraphics[width=\textwidth]{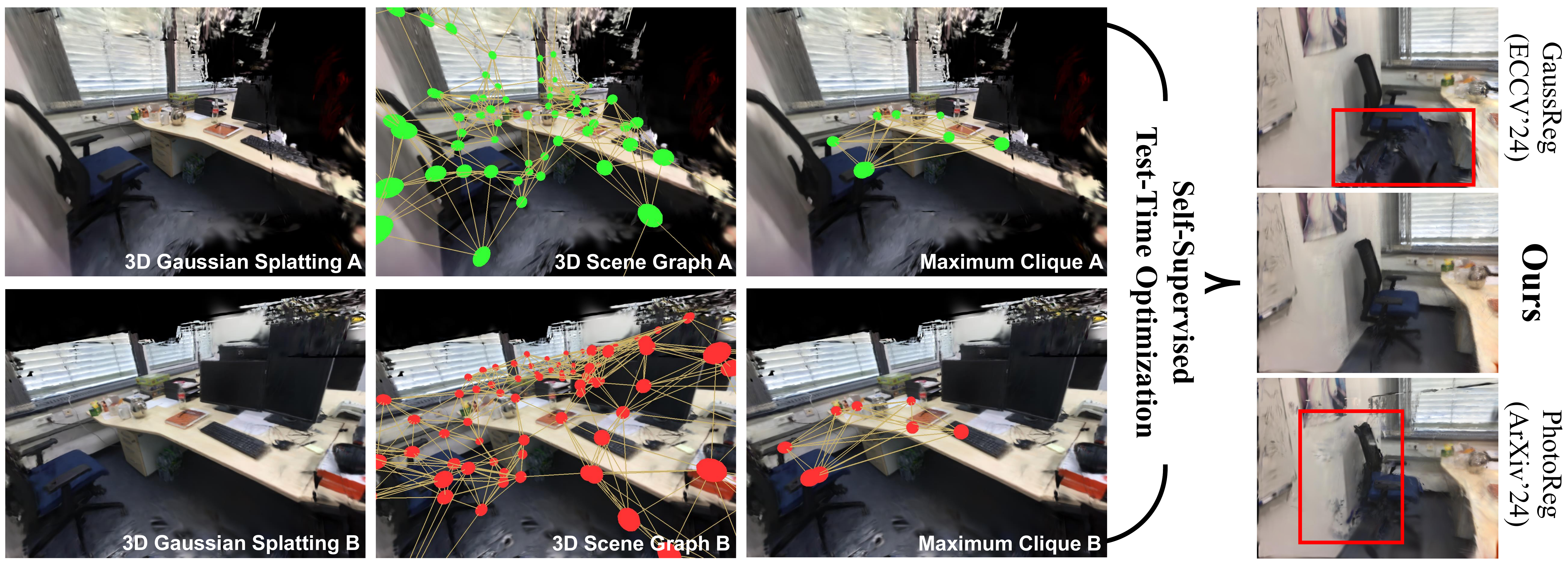}
\caption{\textbf{\ourmodel} performs 3D Gaussian Splatting (3DGS) registration using 3D scene graphs.
Given two partially overlapping 3DGSs, the second column shows the 3D scene graphs constructed from each 3DGS.
In the third column, the two scene graphs are matched, and consistent correspondences are obtained through maximum clique search, without erroneous matches.
}
\vspace{-0.4cm}
\label{figure_1}
\end{figure}

\section{Introduction}
\label{sec:intro}

Large-scale 3D mapping is a fundamental challenge~\cite{cadena2017past} in applications such as robot navigation and AR/VR. Recently, 3D Gaussian Splatting (3DGS)~\cite{kerbl3Dgaussians} has emerged as a new form of map representation, offering fast rendering speed and high visual quality, and is now being explored for large-scale scene representation~\cite{lin2024vastgaussian,liu2024citygaussian}.
However, due to its optimization-based nature that requires refining a massive number of Gaussian primitives, training a single model on large-scale environments leads to excessive memory consumption. This often results in unstable optimization, making it practically infeasible~\cite{papantonakis2024reducing,zhao2024scaling}.
Furthermore, in real-world scenarios, it is often difficult to capture an entire scene at once~\cite{golodetz2018collaborative}.

Consequently, in 3DGS, aligning and merging multiple independently generated scenes is essential for large-scale reconstruction, yet this problem has not been sufficiently explored.
Several approaches have been proposed for 3DGS registration. 
GaussReg~\cite{chang2024gaussreg} is a learning-based framework that performs registration using a point cloud network derived from Gaussian primitives and further refines the alignment through a rendered image-based network. However, it requires training on large 3DGS datasets. 
GaussReg attempts to fuse them based on the distance thresholds between Gaussian points. However, this thresholding-based approach excludes some Gaussians, often resulting in severe hollow regions.
PhotoReg~\cite{yuan2024photoreg} performs coarse alignment of rendered image pairs using DUSt3R~\cite{wang2024dust3r} and refines the alignment via image-based optimization. However, since such off-the-shelf models are not specifically tailored for 3DGS registration, they can introduce potential ambiguities in the initial estimation. This reliance on image-based alignment means that if coarse alignment fails, fine optimization also requires a significant amount of time. Moreover, PhotoReg does not employ a dedicated merging strategy and instead directly combines the scenes, resulting in noticeable artifacts and floaters.

We propose Graph-GSReg, a framework that enables robust alignment and seamless merging of multiple scenes through a 3D scene graph representation~\cite{hughes2022hydra,gu2024conceptgraphs,linok2025beyond}. Although 3DGS offers high-fidelity rendering, its primitives mainly encode low-level attributes such as color, position, and rotation, making them difficult to directly exploit for higher-level reasoning. To address this limitation, we construct a 3D scene graph from a 3DGS and reformulate 3DGS registration as a graph registration problem, enabling consistent alignment across multiple scenes.

To seamlessly merge multiple 3DGS scenes into a unified representation, we further introduce a Self-Supervised Test-Time Optimization (TTO) strategy. Naively merging 3DGS scenes often suffers from occlusion artifacts such as hollows and floaters. To mitigate this, we use the fact that each original scene remains renderable as a self-supervised reference.
Specifically, we render both the merged scene and the corresponding original scene from the same viewpoint, and update the merged Gaussians by minimizing the difference between the two renderings. The renderings from the original scenes serve as reliable references that guide the refinement of the unified representation. This process alleviates artifacts caused by overlapping structures and occlusions while maintaining photometric consistency across the scene.
\cref{figure_1} showcases 3D scene graphs from real-world 3DGS and provides a brief qualitative comparison of our merging results.

The contributions of this paper are summarized as follows:
\begin{itemize}
    \item We propose \ourmodel, a registration framework that constructs a 3D scene graph enriched with geometric and semantic information from 3DGS scenes.
    \item We propose Self-Supervised Test-Time Optimization, a merging strategy that refines aligned 3D Gaussians and yields a seamless and visually consistent scene.
    \item We evaluate our method on both real-world and synthetic benchmarks, achieving competitive registration accuracy and high-quality merged-scene renderings.
\end{itemize}

\section{Related Work}
\subsection{Scene Merging Strategies in Radiance Field}
Several studies have explored aligning and seamlessly merging two scenes in Radiance Field. Based on NeRF~\cite{mildenhall2021nerf}, Block-NeRF~\cite{blocknerf} partitions large-scale scenes into multiple blocks and synthesizes them using inverse distance weighting (IDW) and appearance alignment. 
NeRFuser~\cite{nerfuser} proposes a re-rendering-based registration and a sample-level IDW-Sample blending method to fuse two NeRFs.
In methods based on 3D Gaussian Splatting (3DGS)~\cite{kerbl3Dgaussians}, LoopSplat~\cite{zhu2025loopsplat} registered between loop keyframes detected via NetVLAD~\cite{netvlad} within the SLAM framework, and the additional merging process was performed in the SLAM backend optimization.
GaussReg~\cite{chang2024gaussreg} introduces an end-to-end network for 3DGS scene registration, where the coarse stage aligns scenes using point cloud registration networks~\cite{qin2023geotransformer}, and the fine stage refines them through a rendered image-based fine registration network that extracts volumetric features from overlapping views.
For scene merging in GaussReg, a distance-based thresholding strategy is applied, preserving Gaussians closer to their own scene center. However, this heuristic thresholding approach relies solely on the scene centers, and may erroneously discard important regions when the two scenes differ in size or distribution.
Another 3DGS merging method, PhotoReg~\cite{yuan2024photoreg}, renders all images from each scene, selects the image pair with the highest similarity, and then employs a 3D foundation model such as DUSt3R~\cite{wang2024dust3r} to estimate the initial registration pose.
PhotoReg then refines the alignment via photometric-based optimization for a fixed number of iterations. Although DUSt3R enables image-level registration, failures in the initial stage often affect the subsequent fine optimization, resulting in overall performance degradation.
Moreover, when merging two scenes, PhotoReg simply combines them without any additional processing.

\subsection{Graph-based Representations for Registration}

Graph-based representations are an effective way to transform complex scenes into interpretable forms. In 3D point cloud registration, initial correspondences are modeled as nodes and their geometric compatibility as edges, structurally capturing relations between features. This approach better captures inter-feature dependencies than Euclidean distance–based methods, while the graph connectivity ensures robustness under noise and low overlap by facilitating reliable matching and outlier rejection.
Graph-based registration methods can be divided into learning-based and non-learning-based approaches. Recently, SGAligner~\cite{sarkar2023sgaligner} and SG-PGM~\cite{xie2024sg} have been proposed as learning-based methods. Such approaches leverage graph representations for registration, using GNNs~\cite{brody2021attentive} or Transformer-based models~\cite{fu2021graphmat_tf} to encode structural and semantic relationships that support robust correspondence matching and transformation estimation.
Non-learning-based methods rely on the graph structure directly. Previous studies~\cite{lin2022k_maxcli, zhang2023mac, qiao2023pyramid} extract geometrically consistent inlier correspondences using maximum clique or maximal clique search. Recent methods such as CLIP-Clique~\cite{matsuzaki2024clipclique} enrich nodes with multi-modal features like CLIP~\cite{clip_encoder} embeddings. They select the highest-scoring clique among candidates to perform global localization.
TEASER++~\cite{yang2020teaser} represents correspondences as a graph and performs robust inlier selection, which enables accurate 3D point cloud registration under heavy outlier conditions.
Graph-based approaches are not limited to geometric reasoning. When combined with semantic and visual features, they show significantly improved performance. In this work, we render images from 3D Gaussian Splatting (3DGS) scenes and construct them into a consistent 3D scene graph for registration. The proposed method integrates both semantic and structural information to achieve accurate and robust scene registration.

\section{Method}

\begin{figure*}[t]
  \centering
  \includegraphics[width=0.9\textwidth]{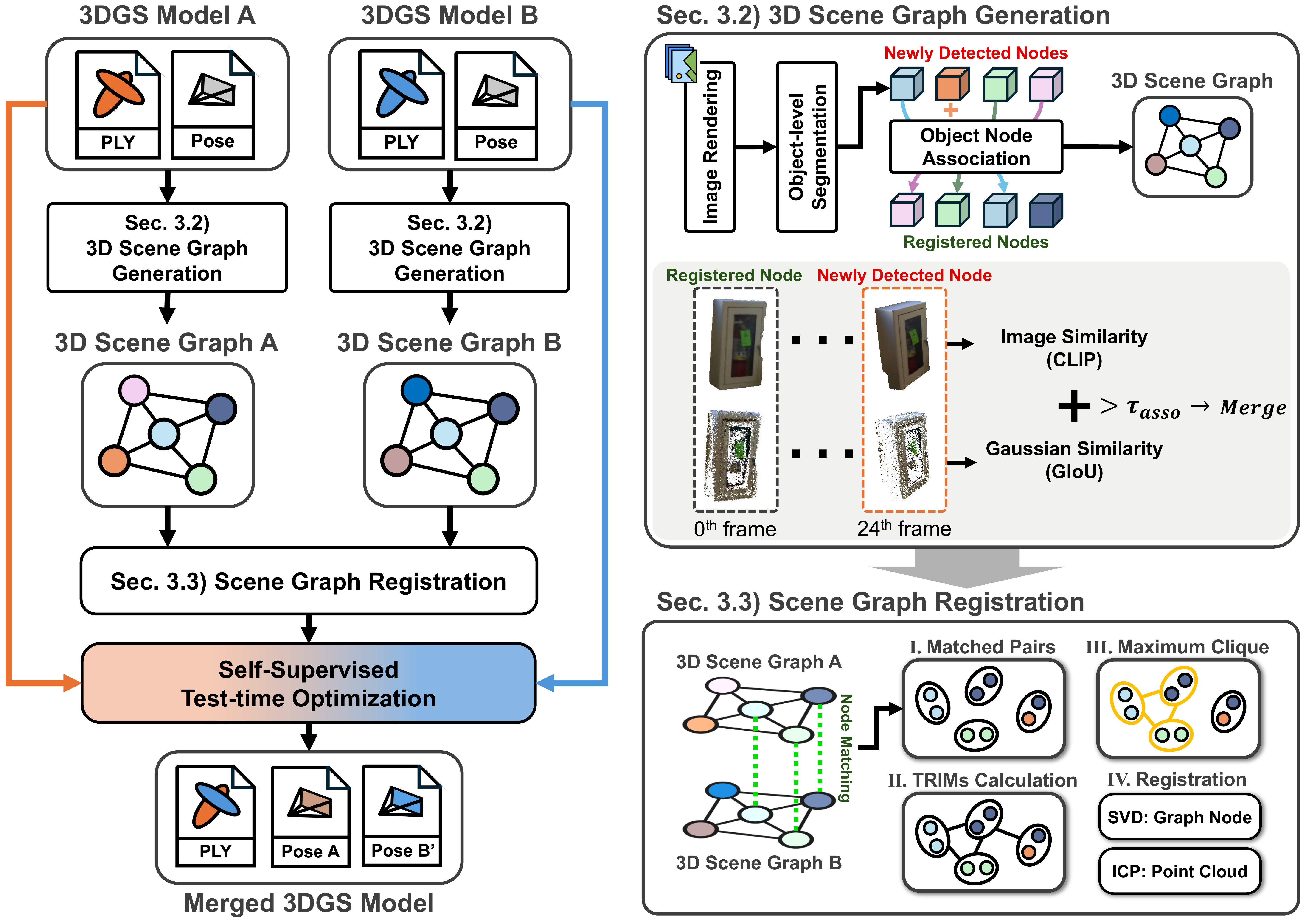}
\caption{\textbf{Overall architecture of \ourmodel}.
Starting from independently constructed 3DGS models, \ourmodel~constructs an object-level 3D scene graph from the 3DGS scenes by associating the same objects across frames to form globally consistent nodes (\cref{3dsg_method}). It then aligns two scenes by matching their scene graphs to build a compatibility graph, where TRIMs-based filtering and clique selection produce reliable correspondences and the final registration (\cref{graph_registration}).}
\label{figure_2}
\end{figure*}

\ourmodel~registers two 3D Gaussian Splatting (3DGS) models, $G^A$ and $G^B$, and seamlessly merges them into a unified 3DGS scene.
Unlike existing 3DGS registration methods that typically rely solely on either geometric or photometric cues, \ourmodel~is designed to convert 3DGS into a 3D scene graph as a higher-level representation. Through this approach, the 3DGS registration problem is reformulated as a graph registration problem. This graph representation jointly encodes rich semantic cues and structural context, enabling robust registration. Finally, to construct a seamless unified scene, we introduce a Self-Supervised Test-Time Optimization that uses the original scenes as references to preserve the visual consistency of the merged scene. 
While we focus on the registration of two models for notational simplicity and clarity, extending this framework to multiple scenes is a natural progression once pairwise registration is established.
\cref{figure_2} illustrates the overall architecture of our proposed method.

\subsection{Problem Setup}
\label{method:problem_setup}
The goal of \ourmodel\ is to merge two unregistered 3D Gaussian Splatting (3DGS) models.
Each scene is defined as
\begin{equation}
S^X = (G^X, C^X), \quad X \in \{A,B\},
\end{equation}
where $G^X$ is the set of Gaussian primitives, and $C^X$ is the set of camera poses.
Each scene $X \in \{A,B\}$ consists of a set of Gaussian primitives
\begin{equation}
    G^X = \{ (\boldsymbol{\mu}_i, \mathbf{\Sigma}_i, \sigma_i, \mathbf{c}_i) \}_{i=1}^{N^X_G},
\end{equation}
where $\boldsymbol{\mu}_i \in \mathbb{R}^3$ is the mean position, 
$\mathbf{\Sigma}_i \in \mathbb{R}^{3 \times 3}$ is the covariance matrix, 
$\sigma_i$ is the opacity, and $\mathbf{c}_i$ is the color attribute of the $i$-th primitive.
Here $N^X_G$ denotes the number of Gaussian primitives in scene $X$.  
The set of camera poses for each scene is defined as $C^X = \{(R_j^X, T_j^X)\}_{j=1}^{N_C^X}$, where $N_C^X$ denotes the total number of image frames in each scene.
The final transformation used to align scene $B$ to scene $A$ is defined as a rigid transformation $\mathbf{T}_B^A = (\mathbf{R}, \mathbf{t}), \ \mathbf{T}_B^A \in SE(3)$, consisting of a rotation matrix $\mathbf{R}$ and a translation vector $\mathbf{t}$. This transformation is applied only to all components of $S^B$.
After applying the transformation, all camera poses of scene $B$ are updated as
\begin{equation}
C^{B'} = \{(\mathbf{R} R_j^B, \mathbf{R} T_j^B + \mathbf{t})\}_{j=1}^{N_C^B}.
\end{equation}
The transformation of the Gaussian primitive set is denoted, for notational simplicity, as ${G^{B}}' = \mathbf{T}_B^A(G^B).$
This represents the result of applying the rigid transformation to all Gaussians in scene $B$.
The transformation of each Gaussian’s position, covariance, and color is performed in the same way as GaussReg~\cite{chang2024gaussreg}. For a robust evaluation, the initial states of the two scenes are set with random orientations and large displacements to ensure no spatial overlap.

\subsection{3D Scene Graph from Gaussian Splatting}
\label{3dsg_method}

Since independently reconstructed 3DGS scenes often contain different Gaussian primitives and inconsistent structures, establishing correspondences directly between Gaussians is unreliable. To address these inconsistencies, we convert each 3DGS scene into a higher-level 3D scene graph representation that jointly captures rich semantic cues and structural information. This globally consistent graph enables accurate registration.

The proposed method constructs a graph utilizing mask-level visual cues and their corresponding Gaussians, obtained from per-frame rendered images of each pre-built 3DGS scene $S^X$, where $X \in \{A,B\}$.
We construct a globally consistent 3D scene graph $\mathcal{G}^X$ for the entire scene by associating nodes corresponding to the same object across multiple frames.
The graph is defined as $\mathcal{G}^X = (\mathcal{V}^X, \mathcal{E}^X)$, consisting of a node set $\mathcal{V}^X$ and an edge set $\mathcal{E}^X$. Since our method constructs the scene graph through cross-frame object association, rendering every frame densely is unnecessary. Accordingly, we render images $I_j^X$ only from a subset of camera poses $C^X_{\text{sampled}}$ uniformly sampled from the entire set.
\begin{equation}
I_j^X = \textit{Render}\!\left(G^X \mid (R_j^X, T_j^X)\right), \quad j \in C^X_{\text{sampled}}, \quad X \in \{A,B\},
\end{equation}
where $\textit{Render}$ denotes the 3DGS rendering procedure.

As shown in \cref{figure_2}, we extract object masks from the rendered images using SAM~\cite{SAM}, utilizing mask information above a certain confidence threshold $c_{mask}$. Denoting the $k$-th specific mask in the $j$-th frame as $m_{j,k}^X$, we first extract the image feature of the corresponding mask region via CLIP~\cite{clip_encoder} to exploit the rich information of the high-fidelity rendered image.
Then, utilizing the current frame's camera pose $(R_j^X, T_j^X)$ and camera intrinsics $\mathbf{K}$, we determine whether the Gaussians in the 3D space belong to the mask.
Specifically, the Gaussian center $\boldsymbol{\mu}_i$ is projected onto the 2D image plane at pixel coordinates $\mathbf{p}_{i,j}^X = [u_{i,j}^X, v_{i,j}^X]^T$ according to
\begin{equation}
z_{i,j}^X
\begin{bmatrix}
\mathbf{p}_{i,j}^X \\
1
\end{bmatrix}
=
\mathbf{K} (R_j^X)^T (\boldsymbol{\mu}_i - T_j^X).
\end{equation}
Here, $z_{i,j}^X$ denotes the depth value along the $Z$-axis in the camera coordinate system. To identify the set of 3D Gaussians corresponding to the object mask $m_{j,k}^X$, we define the subset $\mathcal{A}_{j,k}^X$ consisting of Gaussians whose projected pixel coordinates fall within the mask region and are located in front of the camera ($z_{i,j}^X > 0$)
\begin{equation}
\mathcal{A}_{j,k}^X = \left\{ \boldsymbol{\mu}_i \in G^X \;\middle|\; \mathbf{p}_{i,j}^X \in m_{j,k}^X, \; z_{i,j}^X > 0 \right\}.
\end{equation}
By disregarding attributes such as rotation and scale, the extracted set $\mathcal{A}_{j,k}^X$ can be treated as a 3D point cloud. The mean of the 3D coordinates $\mathbf{x}_{j,k}^X \in \mathbb{R}^3$ in this set is defined as the 3D position of the corresponding object node.
\begin{equation}
\mathbf{x}_{j,k}^X = \frac{1}{|\mathcal{A}_{j,k}^X|} \sum_{\boldsymbol{\mu}_i \in \mathcal{A}_{j,k}^X} \boldsymbol{\mu}_i
\end{equation}

As illustrated in the concrete example in \cref{figure_2}, the image features of the masks and their corresponding Gaussians are used for node association. When compared with nodes appearing in subsequent frames, if the sum of the image feature similarity of mask and the 3D GIoU score between the two Gaussian coordinate sets exceeds the threshold $\tau_{asso}$, the two nodes are considered to represent the same object and are merged. Otherwise, the object is regarded as unseen and added as a new node.
Specifically, during the merge process, the two point clouds are combined and then downsampled to maintain uniform density. Then, we average the two CLIP features and normalize the result.
Finally, to generate as many candidates as possible and identify inliers among them, the nodes are initially constructed as a fully connected graph.

Based on this graph structure, we further enrich each node representation by incorporating structural context.
Appearance-only matching between nodes is often ambiguous when objects share similar visual patterns.
To incorporate structural context, we augment each node’s CLIP feature with a histogram of CLIP features in its 3-hop neighborhood.
While the fully connected graph is used to consider broad candidate relations, the 3-hop neighborhood histogram is computed on a local graph that connects only spatially nearby nodes within a distance threshold. Therefore, this histogram reflects the local context around each object rather than global graph statistics.
The final node embedding is defined by concatenating the normalized CLIP feature with the 3-hop neighborhood CLIP histogram,
\begin{equation}
f(v) = [f_{\text{CLIP}}(v), f_{\text{hist}}(v)],
\quad \forall v \in \mathcal{V}^X,
\end{equation}
which captures both local appearance and the distribution of appearance cues in its neighborhood.

\subsection{Clique-driven Scene Graph Registration}
\label{graph_registration}

Building on this rich information, rather than matching nodes independently, we seek correspondences that can be jointly explained across all nodes. Therefore, we formulate scene graph registration as a global consensus problem rather than a node-wise matching task.
Candidate correspondences are generated using context-enriched node similarity and refined via maximum clique search~\cite{lin2022k_maxcli,yang2020teaser} to identify the largest set of mutually consistent correspondences.

\subsubsection{Compatibility Graph with TRIMs.}

From the constructed two scene graphs $\mathcal{G}^A = (\mathcal{V}^A, \mathcal{E}^A)$ and $\mathcal{G}^B = (\mathcal{V}^B, \mathcal{E}^B)$, we estimate a globally consistent transformation $\mathbf{T}_B^A$ between the scenes. Rather than treating candidate correspondences independently, we construct a compatibility graph $\mathcal{G}_{\text{comp}} = (\mathcal{V}_{\text{comp}}, \mathcal{E}_{\text{comp}})$ that captures pairwise geometric consistency among candidate matches.

Candidate correspondences are generated using node embedding similarity, defined as
\begin{equation}
s(q,p)=\frac{f(q)\cdot f(p)}{\|f(q)\|\|f(p)\|}, \quad
\mathcal{V}_{\text{comp}}=\{(q',p')\mid s(q,p)\ge\tau_{\text{node}},\, q\in\mathcal{V}^A,\, p\in\mathcal{V}^B\}.
\end{equation}

But objects that are visually highly similar, yet located at different spatial positions, can still produce incorrect correspondences. To eliminate such mismatches, we note that for correct correspondences, the distance between any two matched objects in one scene should be consistent with the distance between the corresponding objects in the other scene. Accordingly, we employ TRIMs~\cite{yang2020teaser} to enforce this pairwise distance consistency, and define edges in the compatibility graph as
\begin{equation}
((q'_1,p'_1),(q'_2,p'_2)) \in \mathcal{E}_{\text{comp}}
\iff
1-\tau_{\text{TRIMs}}
<
\frac{\|q'_1-q'_2\|}{\|p'_1-p'_2\|}
<
1+\tau_{\text{TRIMs}} .
\end{equation}

This constraint suppresses geometrically inconsistent matches while preserving correspondences that are mutually compatible under a shared transformation. Consequently, geometrically consistent correspondences become densely connected in the compatibility graph, providing a strong foundation for maximum clique search to identify the largest globally consistent correspondence set.

\subsubsection{Maximum Clique \& Registration.}

Under the TRIMs-based distance invariance constraint, correct correspondences preserve relative distances across scenes and therefore remain mutually connected in the compatibility graph. In contrast, incorrect correspondences may satisfy the distance consistency with some matches but violate it with others. As a result, they fail to form a globally consistent set of correspondences and cannot produce a fully connected structure in the compatibility graph.

Consequently, geometrically consistent correspondences form fully connected subgraphs in the compatibility graph. Based on this observation, we perform a maximum clique search in the compatibility graph.

\begin{figure}[t]
  \centering
  \includegraphics[width=0.8\linewidth]{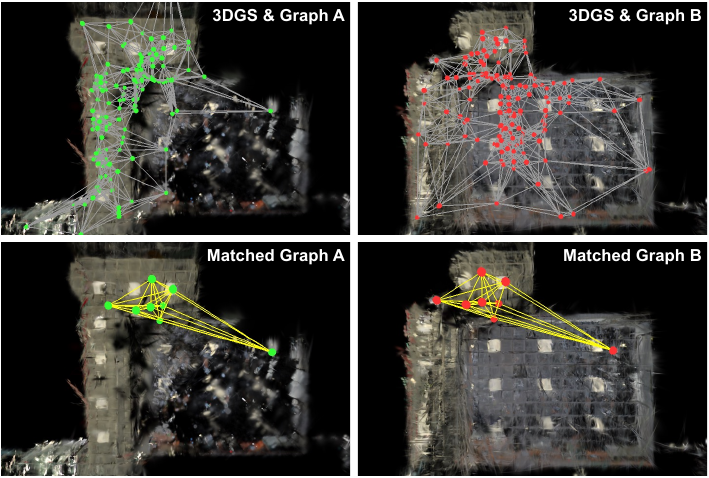}
  \caption{Visualization of the 3D scene graphs from two 3DGS scenes constructed from uHumans2~\cite{uhumans2_kimera}.}
  \label{figure_6}
\end{figure}

\cref{figure_6} shows the node matching process in large-scale indoor environments with partial overlap, where the first row shows the initial nodes obtained from node association and the second row shows the correspondences after TRIMs-based filtering and maximum clique selection. The structural agreement between the two final graphs shows that incorrect matches are effectively removed while preserving geometrically consistent correspondences.

Using the extracted correspondence set, we estimate a rigid transformation $\mathbf{T}_{\text{graph}} \in SE(3)$ using an SVD-based method. When scale estimation is required, the Umeyama algorithm~\cite{umeyama2002least} is applied. The alignment is further refined by performing ICP on the means of the Gaussian primitives initialized with $\mathbf{T}_{\text{graph}}$, yielding the final transformation $\mathbf{T}_B^A \in SE(3)$.

\subsection{Self-Supervised Test-Time Optimization for Seamless Merging}
\label{merging_opti}

The primary purpose of \ourmodel~is to seamlessly fuse independently reconstructed scenes without information loss or visual degradation. 
However, inherent imperfections in rigid registration often lead to overlapping artifacts, occlusions, and redundant densities when naively aligning two Gaussian sets.

To address artifacts, occlusions, and redundant densities caused by inherent imperfections in 3D registration, we introduce a Self-Supervised Test-Time Optimization (TTO) strategy. Instead of relying on heuristic pruning~\cite{chang2024gaussreg}, our core insight is to leverage the original scenes as reliable rendering references. Although the ground-truth for the unified scene is absent, each individual scene, $S^A$ and the transformed $S^{B'}$, is still renderable from its original viewpoints. We utilize these source renderings as references to guide the refinement of the unified representation.
Specifically, after alignment, we first construct a naive union of the two Gaussian sets, $G^A \cup G^{B'}$. To suppress redundant Gaussians in the partially overlapping regions and establish a memory-efficient, stable initialization, we apply a spatial voxelization with a resolution of 0.01\,m to this union. Let $G^{\text{merged}}$ denote the Gaussian parameters of this initial merged scene, and let $C^{\text{merged}} = C^A \cup C^{B'}$ denote the combined set of all camera poses.

During the TTO phase, we refine the merged scene using renderings from the original scenes as references. From the same viewpoint, we first render the original scene to obtain a reference image $I_j^X = \textit{Render}(G^X \mid R_j^X, T_j^X)$, and then render the merged scene using $G^{\text{merged}}$ from the same viewpoint. The merged Gaussians are updated by minimizing the photometric discrepancy between the two renderings.
Formally, the merged scene is optimized by minimizing the following $L_1$ rendering loss
\begin{equation}
\arg\min_{G^{\text{merged}}}
\sum_{j}
\left\|
\textit{Render}(G^{\text{merged}} \mid R_j^X, T_j^X) - I_j^X
\right\|_1,
\quad X \in \{A, B'\}.
\end{equation}

Unlike geometric thresholding based solely on distance~\cite{chang2024gaussreg}, this self-supervised mechanism is \textit{based on the expectation that the merged scene should preserve the full content of the original scenes.} Under this assumption, it naturally mitigates occlusion artifacts and maintains photometric consistency across the global scene. Moreover, this refinement achieves strong performance with only one minute of optimization and produces a visually seamless and complete scene representation.

\section{Experiments}

\subsection{Experimental Setup}

In this work, we evaluate both registration and merging performance on the real-world dataset ScanNet-GSReg~\cite{chang2024gaussreg} and the synthetic dataset uHumans2~\cite{uhumans2_kimera}. For registration performance (\cref{registration_eval}), we report Relative Rotation Error (RRE), Relative Translation Error (RTE), Relative Scale Error (RSE), and Absolute Translation Error (ATE). For merging performance (\cref{merging_eval}), we compare rendered and ground-truth images using PSNR, SSIM~\cite{ssim}, and LPIPS~\cite{lpips}.

To evaluate performance in larger indoor environments, we conduct additional experiments on a dataset constructed from uHumans2. Specifically, we generate 69 scene pairs from four static RGB-D sequences without dynamic objects. Each 3DGS model is reconstructed from 150 frames obtained by temporally subsampling the RGB-D sequence at 5 fps, and each scene pair contains 51 overlapping frames.
All comparisons are performed by perturbing the initial transformation with random noise, and the reported metrics represent the mean error across all pairs. All experiments are conducted on an NVIDIA RTX 4070 and an Intel Core i5-13600.

Our method requires a preprocessing step that re-renders a pre-built 3DGS
to generate a scene graph. However, this step is lightweight compared to the
training cost of learning-based approaches and is performed only once per scene.
The constructed graph can be reused across multiple registration tasks.

Additional implementation details, extended analyses, and qualitative results are provided in the supplementary material.

\begin{table}[t]
\centering
\caption{Quantitative comparison of Gaussian Splatting registration results on the ScanNet-GSReg~\cite{chang2024gaussreg} dataset. (* indicates values reported in the original papers.)}
\label{tab:scannetreg_reg}

\scriptsize
\setlength{\tabcolsep}{3pt}
\begin{tabular*}{0.8\linewidth}{l @{\hspace{1.1cm}} c@{\hspace{0.35cm}}c@{\hspace{0.35cm}}c@{\hspace{0.35cm}}c}
\toprule
Method & RRE ($^\circ$) $\downarrow$ & RTE $\downarrow$ & RSE $\downarrow$ & Time (s) $\downarrow$ \\
\midrule
GaussReg* (Coarse)~\cite{chang2024gaussreg}  & 3.403 & 0.061 & 0.034 & 3.700    \\
GaussReg* (w/ Fine)~\cite{chang2024gaussreg} & 2.827 & 0.042 & 0.032 & 4.800    \\
PhotoReg (Coarse)~\cite{yuan2024photoreg} & 7.825 & 0.082 & - & 16.231    \\
PhotoReg (w/ Fine)~\cite{yuan2024photoreg} & 7.366 & 0.072 & - & 585.216    \\
Liu et al.*~\cite{liu2025automated} & 2.595 & 0.045 & \textbf{0.013} & 5.041    \\  
\midrule
Ours (Coarse)  & 3.247 & 0.039 & \textbf{0.013} & \textbf{2.784}   \\
Ours (w/ Fine) & \textbf{1.970} & \textbf{0.025} & - & 3.658  \\
\bottomrule
\end{tabular*}
\end{table}

\begin{table}[t]
\centering
\caption{Quantitative comparison of Gaussian Splatting registration results on the uHumans2~\cite{uhumans2_kimera} dataset.}
\label{tab:uhumans2}

\scriptsize
\setlength{\tabcolsep}{3pt}

\begin{tabular*}{0.8\linewidth}{l @{\hspace{1.0cm}} cccc}
\toprule
Method 
& RRE ($^\circ$) $\downarrow$ 
& RTE $\downarrow$
& ATE (m) $\downarrow$
& Time (s) $\downarrow$ \\
\midrule
TEASER++ \& ICP~\cite{yang2020teaser} 
& 9.099 & 0.071 & 1.669 & 4.570 \\
MAC (FPFH)~\cite{zhang2023mac} & 108.089 & 0.652 & 17.118 & 6.359 \\
MAC (FCGF)~\cite{zhang2023mac} & 60.021 & 0.335 & 8.862 & 5.462 \\
\midrule
PhotoReg (Coarse)~\cite{yuan2024photoreg} & 15.895 & 0.154 & 6.916 & 17.814 \\
PhotoReg (w/ Fine)~\cite{yuan2024photoreg} & 9.959 & 0.077 & 2.754 & 342.754 \\
Ours (Coarse) & 1.748 & 0.010 & 0.243 & \textbf{1.805} \\
Ours (w/ Fine) & \textbf{0.711} & \textbf{0.006} & \textbf{0.192} & 3.782 \\
\bottomrule
\end{tabular*}
\end{table}


\subsection{3DGS Registration Evaluation}
\label{registration_eval}

As reported in \cref{tab:scannetreg_reg}, we evaluate 82 partially overlapping scene pairs from ScanNet-GSReg~\cite{chang2024gaussreg}. GaussReg~\cite{chang2024gaussreg} performs both coarse and fine registration using supervised training and requires extensive datasets of completed 3DGS models. PhotoReg~\cite{yuan2024photoreg} does not require training, but it relies only on images. Its coarse registration based on a 3D foundation model~\cite{wang2024dust3r} produces large errors in some scenes, which leads to poor fine alignment. Liu et al.~\cite{liu2025automated} merge 3DGS using skeleton structures and show strong performance among recent methods.
Our 3D scene graph-based registration achieves the best final performance, with a clear improvement in RTE. We do not report RSE for PhotoReg, as it estimates the scale of the two scenes separately, or for our ICP-based refinement, which is restricted to SE(3).

We further evaluate registration on the uHumans2~\cite{uhumans2_kimera} dataset using 69 scene pairs. We focus on methods applicable without large-scale supervised training, including TEASER++~\cite{yang2020teaser}, MAC~\cite{zhang2023mac}, and PhotoReg~\cite{yuan2024photoreg}. TEASER++ extracts FPFH~\cite{rusu2009fast} features for global registration and then applies ICP for refinement. Meanwhile, MAC constructs compatibility graphs based on features extracted by FPFH or FCGF~\cite{FCGF2019} and solves the registration problem through maximal clique search to identify the most consistent set of correspondences. However, these traditional point cloud-based methods often face challenges in noisy environments where incorrect feature matches lead to coarse alignment failures. In contrast, our 3D scene graph-based registration with ICP refinement achieves stable and high accuracy across all scenes. Detailed runtime analysis is reported in the supplementary material.

\subsection{3DGS Merging Quality Evaluation}
\label{merging_eval}

In this subsection, we compare the rendered image quality of different 3DGS merging methods. The evaluation is based on comparing rendered images of the merged scenes with ground truth images. The Oracle represents an upper bound as it renders each scene independently before merging and compares them with the ground truth. Since no merging is performed, this setting is free from duplication, removal, or artifacts introduced during the merging process.

As shown in \cref{tab:merging_all_dataset}, we apply all merging methods to the same fixed registration results produced by \ourmodel~on ScanNet-GSReg and uHumans2, isolating the effect of the merging strategy. Under this setting, our Self-Supervised Test-Time Optimization-based merging achieves the best image quality. PhotoReg simply merges the aligned scenes without further processing, while GaussReg selects Gaussians based on distance thresholds. 

We further evaluate merging under ground-truth alignment and observe the same trend, indicating that the gain comes from the merging process itself rather than relying on registration accuracy.

\begin{figure*}[t]
  \centering
  \includegraphics[width=\textwidth]{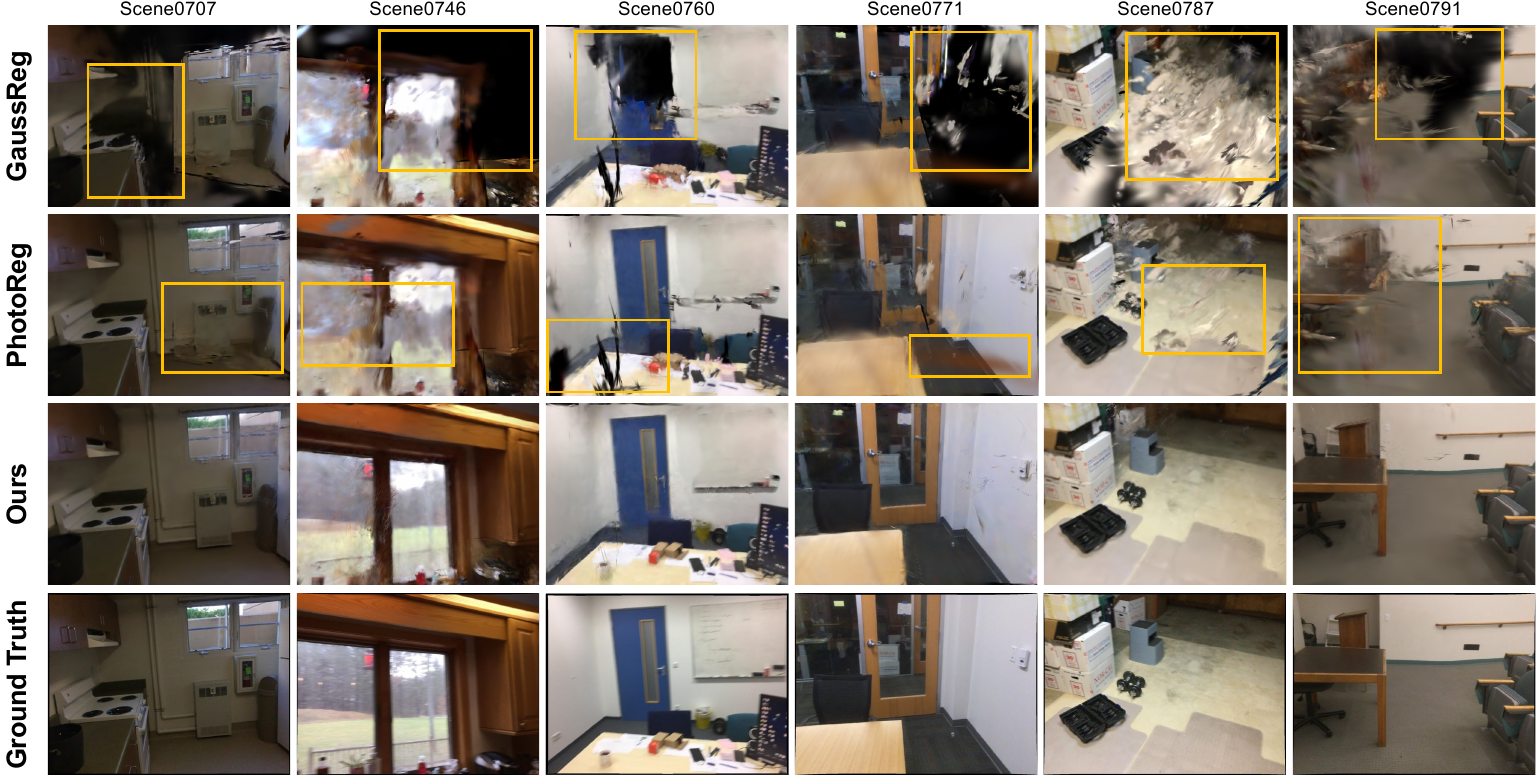}
\caption{
Qualitative results on the ScanNet-GSReg~\cite{chang2024gaussreg} dataset. Each row shows rendered images from merged Gaussians. The yellow box highlights regions with severe hollows or occlusions.
}
\label{figure_4}
\end{figure*}

\begin{figure}[t]
  \centering
  \includegraphics[width=1.0\linewidth]{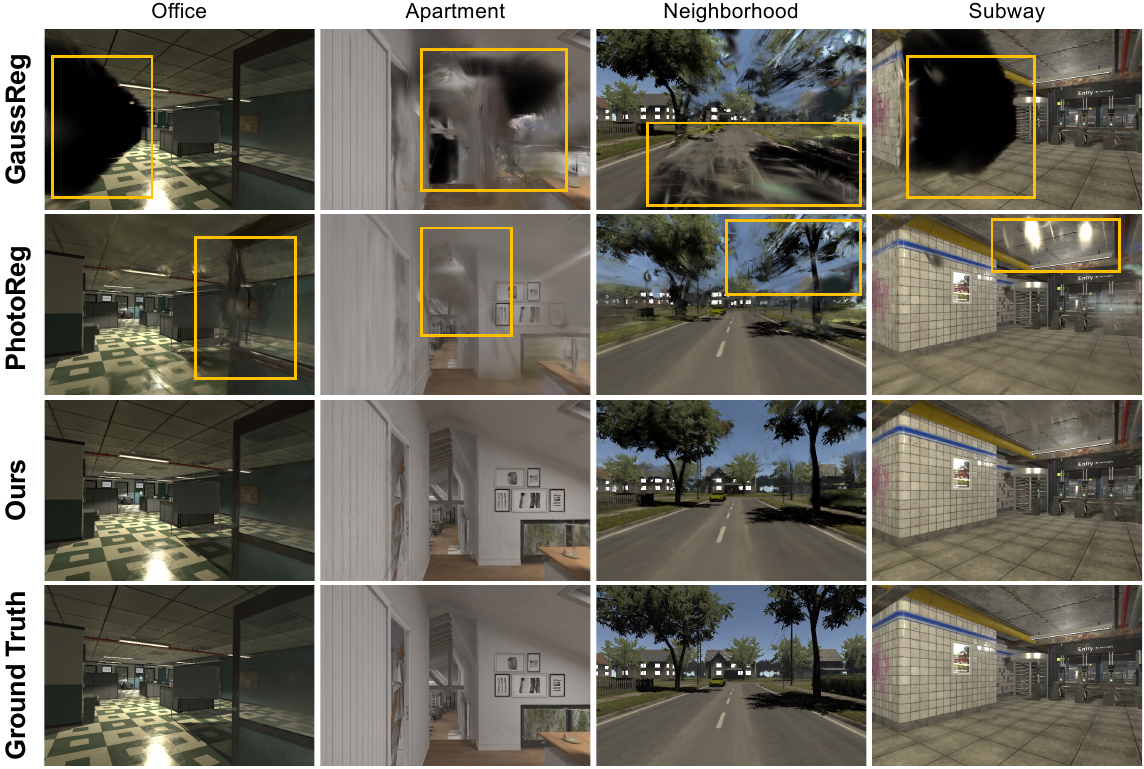}
\caption{
Qualitative results on the uHumans2~\cite{uhumans2_kimera} synthetic dataset. Each row shows rendered images from merged Gaussians. The yellow box highlights regions with severe hollows or occlusions.
}
\label{figure_5_uhumans}
\end{figure}

\cref{figure_4} and~\cref{figure_5_uhumans} show qualitative results on ScanNet-GSReg and uHumans2 using the same fixed registration results. GaussReg often produces hollow regions or strong artifacts due to its distance-based selection. PhotoReg accumulates redundant Gaussians, which degrades the scene quality. In contrast, our method effectively reduces incompleteness and floater artifacts.

\begin{table}[t]
\centering
\caption{Quantitative comparison of Gaussian Splatting merging results on the ScanNet-GSReg~\cite{chang2024gaussreg} and uHumans2~\cite{uhumans2_kimera} datasets.}
\label{tab:merging_all_dataset}

\footnotesize
\setlength{\tabcolsep}{3.5pt}

\resizebox{0.8\columnwidth}{!}{%
\begin{tabular}{l l ccc ccc}
\toprule
& & \multicolumn{3}{c}{ScanNet-GSReg~\cite{chang2024gaussreg}} & \multicolumn{3}{c}{uHumans2~\cite{uhumans2_kimera}} \\
\cmidrule(lr){3-5} \cmidrule(lr){6-8}
Registration & Merging 
& PSNR $\uparrow$ & SSIM $\uparrow$ & LPIPS $\downarrow$
& PSNR $\uparrow$ & SSIM $\uparrow$ & LPIPS $\downarrow$ \\
\midrule

\multicolumn{2}{c}{\textbf{Oracle}} 
& 22.2913 & 0.8566 & 0.3347
& 32.0999 & 0.9218 & 0.1416 \\
\midrule

\multirow{3}{*}{Ground Truth}
& PhotoReg~\cite{yuan2024photoreg}
& 20.1588 & 0.8117 & 0.3844
& 20.9777 & 0.7081 & 0.3316 \\

& GaussReg~\cite{chang2024gaussreg}
& 20.9512 & 0.8306 & 0.3551
& 23.3410 & 0.7368 & 0.2871 \\

& \ourmodel
& \textbf{21.8954} & \textbf{0.8465} & \textbf{0.3484}
& \textbf{25.4032} & \textbf{0.7608} & \textbf{0.2801} \\

\midrule

\multirow{3}{*}{\ourmodel}
& PhotoReg~\cite{yuan2024photoreg}
& 19.3298 & 0.7926 & 0.4026
& 20.7897 & 0.7048 & 0.3370 \\

& GaussReg~\cite{chang2024gaussreg}
& 20.1133 & 0.8089 & 0.3715
& 23.1519 & 0.7321 & 0.2904 \\

& \ourmodel
& \textbf{21.5248} & \textbf{0.8304} & \textbf{0.3698}
& \textbf{25.2425} & \textbf{0.7607} & \textbf{0.2868} \\

\bottomrule
\end{tabular}
}
\end{table}

\subsection{Ablation Study}

\subsubsection{3D Gaussian Splatting Registration.} To evaluate the effectiveness of each component, we conducted ablation studies on the ScanNet-GSReg~\cite{chang2024gaussreg} dataset, and the results are summarized in \cref{tab:scannet_ablation}. Incorporating a CLIP histogram improved registration performance compared to not using it.
When TRIMs were not applied and all nodes were used, the accuracy decreased, confirming the necessity of geometrically validated node selection. Moreover, skipping clique search after TRIMs in the compatibility graph also led to degraded performance.
Adding FPFH~\cite{rusu2009fast} features improved accuracy but increased computation time, so we exclude it from the final design, though it may be useful when point clouds are consistent or higher precision is needed. Finally, compared to CLIP-Clique~\cite{matsuzaki2024clipclique}, which selects the maximal clique with the highest CLIP similarity among nodes, our maximum clique strategy consistently achieved superior performance by leveraging more nodes. Overall, these results demonstrate that each component contributes to balancing registration accuracy and computational efficiency.

\vspace{-0.2cm}

\begin{table}[t]
\centering

\begin{minipage}[t]{0.48\textwidth}
\centering
\caption{Ablation study of the 3D scene graph registration method on the ScanNet-GSReg~\cite{chang2024gaussreg} dataset.}
\label{tab:scannet_ablation}
\resizebox{\linewidth}{!}{%
\begin{tabular}{lcccc}
\toprule
Method & RRE $\downarrow$ & RTE $\downarrow$ & RSE $\downarrow$ & Time (s) $\downarrow$  \\
\midrule
w/o CLIP Histogram & 4.865 & 0.052 & 0.019 & \textbf{2.755} \\
w/o TRIMs & 6.047 & 0.140 & 0.104 & 2.903 \\
w/o Maximum Clique & 6.575 & 0.151 & 0.120 & 2.801 \\
w/ FPFH feature & \textbf{3.190} & \textbf{0.039} & \textbf{0.013} & 3.645 \\
\midrule
CLIP Histogram-5 & 3.867 & 0.045 & 0.015 & 4.341 \\
Maximal Clique (Best CLIP Score)~\cite{matsuzaki2024clipclique} & 3.981 & 0.045 & \textbf{0.013} & 2.805 \\
\textbf{Ours} & 3.247 & \textbf{0.039} & \textbf{0.013} & 2.784 \\
\bottomrule
\end{tabular}
}
\end{minipage}
\hfill
\begin{minipage}[t]{0.48\textwidth}
\centering
\caption{
Ablation study and memory usage comparison of the 3DGS merging method on the ScanNet-GSReg~\cite{chang2024gaussreg} dataset.
}
\label{tab:ablation_adaptive_voxel}
\begin{threeparttable}
\resizebox{\linewidth}{!}{%
\begin{tabular}{l c c c c}
\toprule
Setting & PSNR $\uparrow$ & SSIM $\uparrow$ & LPIPS $\downarrow$ & Size (MB) $\downarrow$\\
\midrule
\textbf{Oracle} & 22.2913 & 0.8566 & 0.3347 & - \\
\midrule
GaussReg~\cite{chang2024gaussreg} & 20.1905 & 0.8106 & 0.3698 & 148.18 \\
Ours (w/o Voxelization) & 21.4538 & 0.8304 & \textbf{0.3675} & 238.10 \\
Ours (w/ Voxelization) & \textbf{21.5248} & \textbf{0.8304} & 0.3698 & \textbf{82.84} \\
\bottomrule
\end{tabular}
}
\end{threeparttable}
\end{minipage}
\vspace{-0.2cm}

\end{table}

\subsubsection{3D Gaussian Splatting Merging.}

As discussed in \cref{merging_opti}, we also perform voxelization after registration before merging the two scenes. 
If the scenes are simply merged without any additional processing, a large number of duplicated Gaussians remain in the overlapped regions, which leads to a significant increase in storage size. 
As shown in \cref{tab:ablation_adaptive_voxel}, such a simple merging setting requires much larger memory compared to GaussReg, which reduces storage by selecting Gaussians based on distance. 
In contrast, our method applies voxelization before merging, effectively eliminating redundant Gaussians and achieving the smallest storage size, while maintaining or even improving merging performance.

\section{Limitations}
Our method requires a preprocessing step to construct a scene graph from each 3DGS scene prior to registration. This step involves image rendering and object-level information extraction, which introduces additional computation time. However, the graph is generated only once per scene and can be reused for multiple registration tasks. Moreover, this design removes the need for large-scale training data and can be applied regardless of environment or domain as long as a 3DGS scene is available.

\section{Conclusion}
We propose \ourmodel, a framework for robust registration and seamless merging of 3D Gaussian Splatting (3DGS) scenes. By leveraging 3D scene graphs, we reformulate registration as a graph matching problem and achieve consistent alignment without training. In addition, our Self-Supervised Test-Time Optimization utilizes the existing Gaussian scenes to update merged Gaussians, alleviating occlusions and scene incompleteness and producing seamless merged results. Overall, \ourmodel~provides an effective solution for large-scale 3DGS-based scene reconstruction and long-term map management in robotics and 3D mapping applications.

\newpage

\bibliographystyle{splncs04}
\bibliography{main}



\section*{Supplementary material}

# Supplementary Material of Graph-GSReg: Leveraging 3D Scene Graphs for Gaussian Splatting Registration

Jaewon Lee[1], Mangyu Kong[1], and Euntai Kim[1,2]⋆

[1] Yonsei University, Seoul, Republic of Korea
[2] Korea Institute of Science and Technology, Seoul, Republic of Korea
{leejaewon,mangyu0929,etkim}@yonsei.ac.kr
https://lee-jaewon.github.io/Graph-GSReg/

We provide additional details and analyses of Graph-GSReg in this supplementary material. Specifically, we include additional qualitative comparisons, multi-scene registration results, an analysis of Test-Time Optimization, details of dataset construction, ablation studies, failure mode analysis, implementation details, and additional evaluations in challenging outdoor environments.

## 1 Qualitative Comparison of Registration Results

The main paper presents a 3DGS merging quality evaluation, demonstrating the effectiveness of our Self-Supervised Test-Time Optimization (TTO) independent of the underlying registration results.

To further analyze the robustness of the registration performance itself, Fig. 1 shows rendered scenes based on the registration results before applying TTO. We compare our method with GaussReg [2] (Coarse) and PhotoReg [9]. For all methods, the images are rendered directly after registration without applying any merging strategy.

Naively merging two scenes without additional refinement inevitably produces visual artifacts such as hollows and floaters due to occlusions between the two scenes. Nevertheless, our method achieves more accurate registration from the outset and produces cleaner rendering quality than GaussReg and PhotoReg even without additional refinement. In particular, as shown in Fig. 1, inaccurate registration leads to severe artifacts, such as duplicated objects, scenes occluded by large floaters. Such accurate initial alignment provided by our method offers a strong foundation for the subsequent TTO stage to further refine the merged scene.

One potential issue is that when the registration error is large, a greater number of Gaussians must be corrected during optimization, making it difficult to achieve sufficient improvement within the same optimization time. Therefore, the higher initial alignment accuracy of our method enables TTO to operate more efficiently, ultimately producing more stable and higher-quality merged scenes.

⋆ Corresponding author

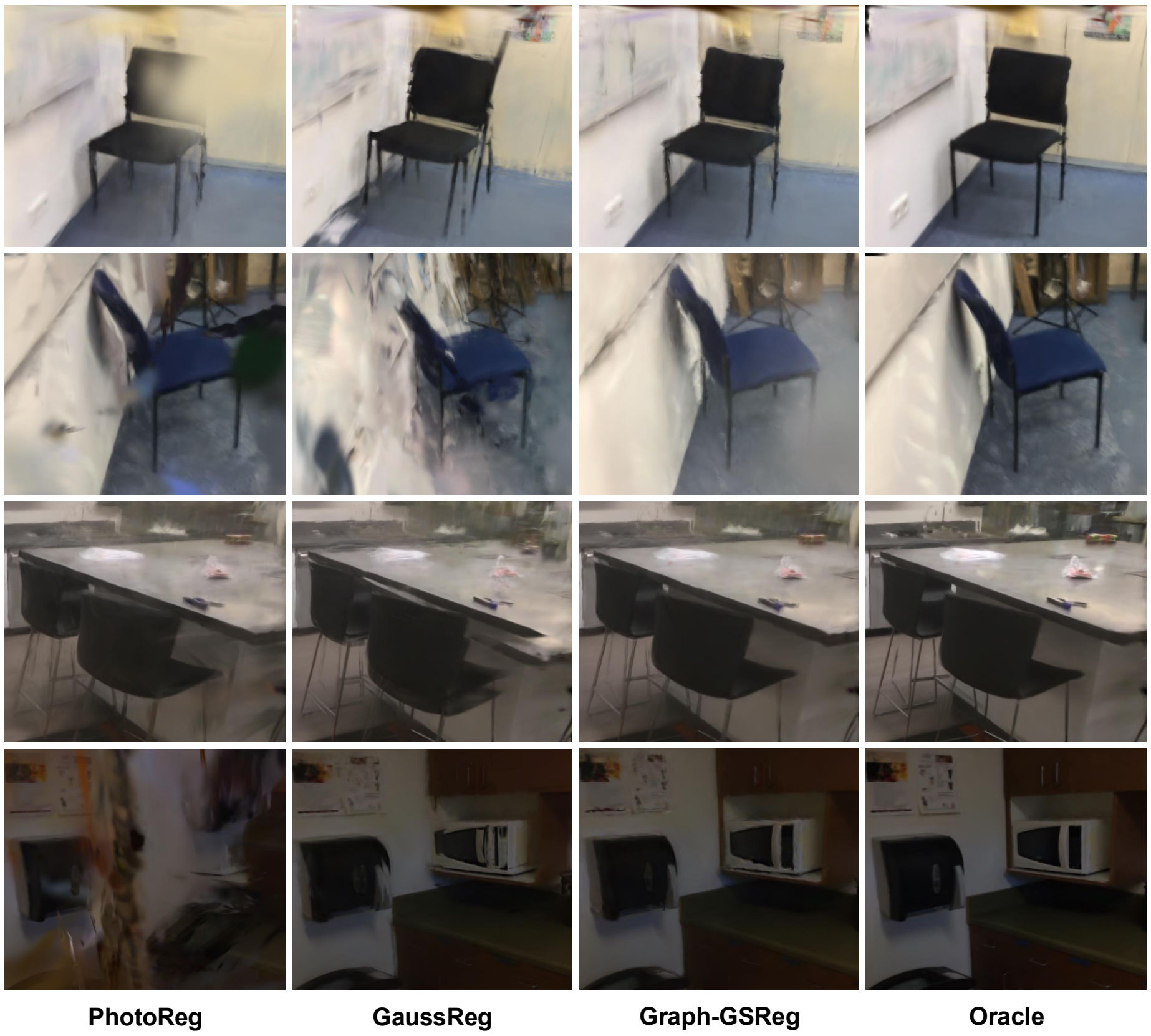


**Fig. 1: Comparison without Test-Time Optimization.** Rendered image of registration results on ScanNet-GSReg [2] before seamless merging. Oracle shows renderings of the original, unmerged 3DGS scenes at the same camera poses. Graph-GSReg already produces clean renderings even before our TTO refinement.

## 2 Feasibility of Multi-Scene Registration

For notational simplicity, Graph-GSReg is described for pairwise scene registration. However, if pairwise registration is feasible, the proposed method naturally extends to multi-scene registration. To demonstrate this scalability, Fig. 2 presents qualitative results on the office sequence of the large-scale synthetic dataset uHumans2 [7], where multiple scenes are sequentially registered to form a unified scene.

The pipeline based on TEASER++ [8] for global estimation followed by ICP [1] refinement achieves high registration accuracy for most scene pairs but fails for several pairs, resulting in misaligned scenes. In contrast, the proposed Graph-GSReg consistently produces accurate alignments across all scenes by

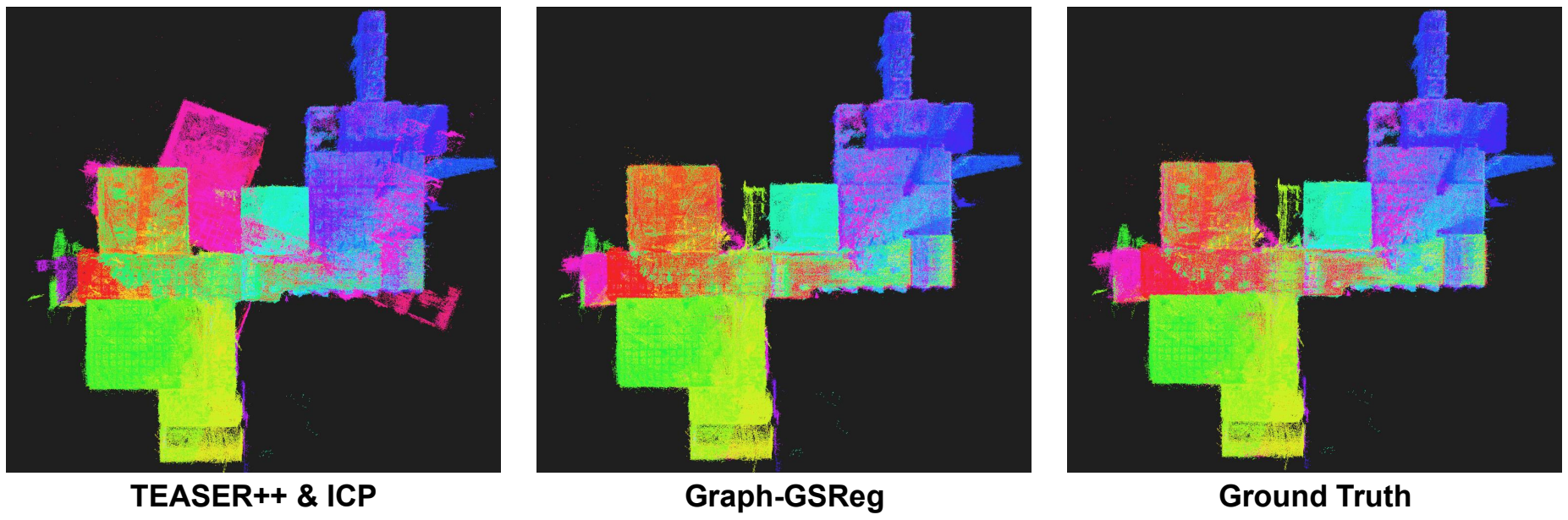


**Fig. 2: Qualitative results of multiple-scene registration.** Alignment of 20 individual 3DGS scenes from the office sequence of uHumans2 [7]. The TEASER++ & ICP pipeline shows incorrect alignments in several scenes, whereas the proposed method produces results nearly identical to the ground truth. Different colors represent different individual scenes.

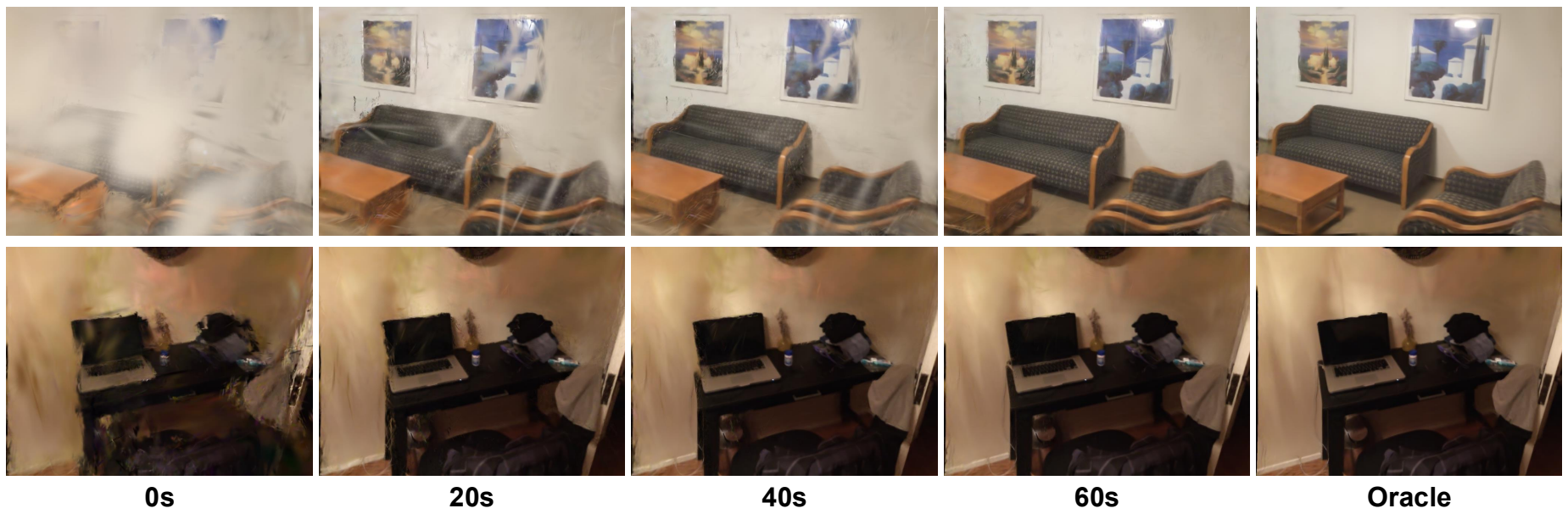


**Fig. 3: Qualitative results on ScanNet-GSReg [2] showing how the optimization evolves over time.** When the two Gaussian scenes are naively merged after registration, severe occlusion artifacts appear, as shown at 0s. After only 20 seconds of optimization, many of these occlusions are resolved, and the scene progressively approaches the appearance of the oracle rendering.

leveraging scene graph matching, yielding results that are nearly identical to the ground truth.

## 3 Analysis of Self-Supervised Test-Time Optimization

In this section, we present a detailed qualitative evaluation and analysis of the proposed Self-Supervised Test-Time Optimization.

Fig. 3 illustrates how the optimization evolves over time. The occlusions introduced by naively merging two scenes are gradually resolved as the optimization proceeds, and the merged scene becomes increasingly similar to the oracle rendering. Here, the oracle is obtained by simply rendering each original 3DGS scene independently and represents the upper bound that the merged scene can ideally achieve.

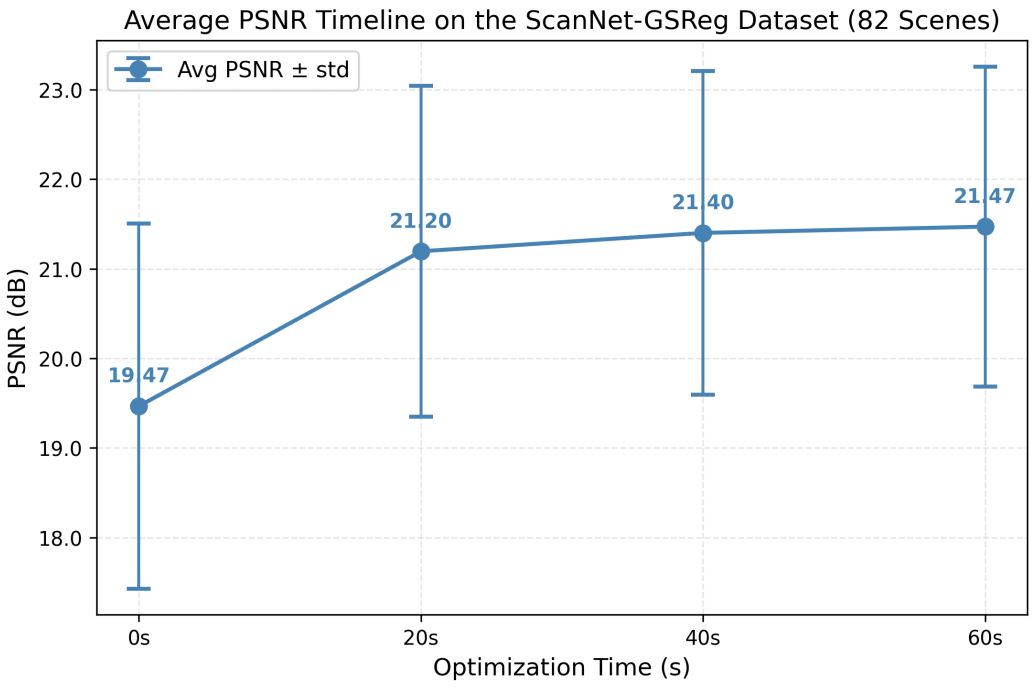


PSNR Improvement Table

| Interval | ΔdB | Δ% | epochs |
|---|---|---|---|
| 0s→20s | +1.7286 dB | +8.88% | 2.21 |
| 20s→40s | +0.2052 dB | +0.97% | 4.47 |
| 40s→60s | +0.0699 dB | +0.33% | 6.76 |
| **0s→60s (total)** | **+2.0038 dB** | **+10.29%** | **6.76** |

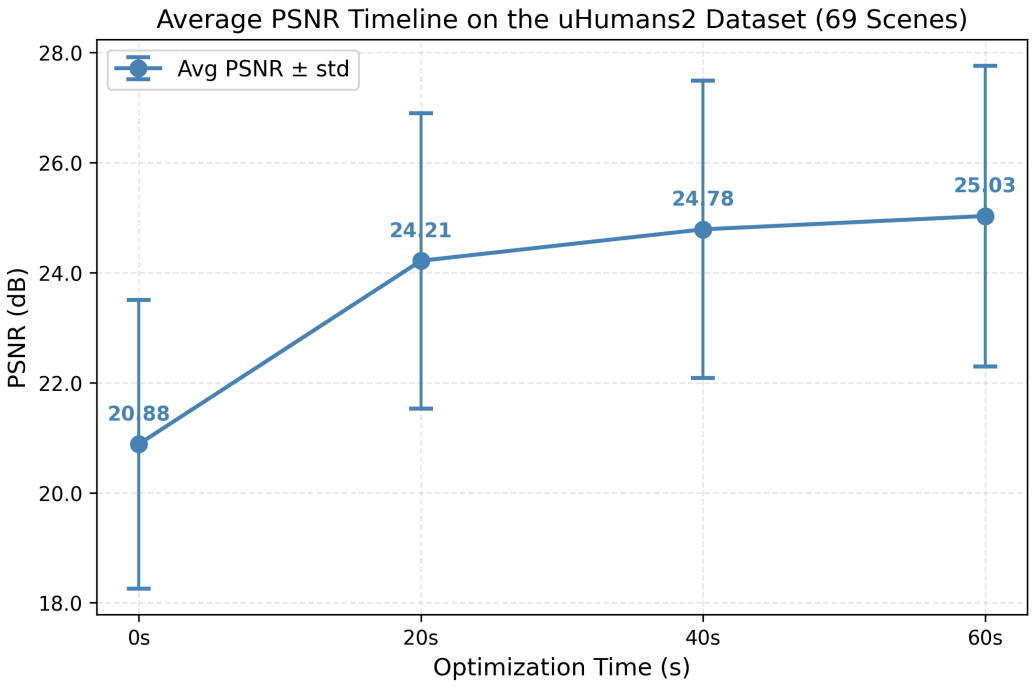


PSNR Improvement Table

| Interval | ΔdB | Δ% | epochs |
|---|---|---|---|
| 0s→20s | +3.3305 dB | +15.95% | 1.06 |
| 20s→40s | +0.5699 dB | +2.35% | 2.13 |
| 40s→60s | +0.2427 dB | +0.98% | 3.21 |
| **0s→60s (total)** | **+4.1432 dB** | **+19.84%** | **3.21** |

**Fig. 4: PSNR improvement during Test-Time Optimization.** Top: Average PSNR trend over 82 test scenes from ScanNet-GSReg [2]. Bottom: Average PSNR trend over 69 merged scenes from the uHumans2 dataset [7].

For a more quantitative analysis of the optimization behavior, Fig. 4 reports the average PSNR improvement on 82 merged scenes from ScanNet-GSReg [2] and 69 merged scenes from the uHumans2 [7] dataset.

In ScanNet-GSReg, most scenes exhibit an increase of approximately 9% within the first 20 seconds, by which point the scene is already close to an almost optimal state. Afterward, the improvement continues more gradually. This early improvement corresponds to roughly two passes over the full camera set in ScanNet-GSReg (i.e., about two epochs), indicating that repeatedly observing the same views quickly benefits the optimization.

Furthermore, a similarly large performance improvement is observed on uHumans2, where the average PSNR increases by more than 15% within the first 20 seconds. Since each scene in uHumans2 contains more images, the optimization observes the full camera set only about once within this 20 seconds (i.e., roughly one epoch). After that, the performance continues to improve steadily, and the PSNR can increase by as much as 4 dB.

Since the proposed TTO process does not rely on ground-truth images but instead uses the original scenes to guide the merged scene toward a visually consistent result, PSNR can be evaluated at any time during the optimization.

**Table 1: Comparison of node feature representations for graph-based scene registration on ScanNet-GSReg [2].** CLIP features provide more stable node matching than DINOv2, and incorporating the proposed 3-hop histogram further improves the registration accuracy.

| CLIP [6] | DINOv2 [4] | 3-Hop Histogram | RRE (°) ↓ | RTE ↓ | RSE ↓ | Success Rate ↑ |
|---|---|---|---|---|---|---|
| | ✓ | | 10.782 | 0.149 | 0.022 | 0.524 |
| | ✓ | ✓ | 12.328 | 0.164 | 0.044 | 0.549 |
| ✓ | | | 4.865 | 0.052 | 0.019 | **1.0** |
| ✓ | | ✓ | **3.247** | **0.039** | **0.013** | **1.0** |

Depending on the user's requirements, the TTO process can be flexibly terminated once a desired PSNR level or a predefined PSNR improvement is reached, which provides a practical advantage in real-world applications.

Additional qualitative results are presented in Fig. 6, Fig. 7 and Fig. 8.

## 4 Analysis of Node Feature Representations

We use CLIP [6] features as the node representation in the scene graph. Recently, self-supervised visual representations such as DINOv2 [4] have shown strong performance across various visual understanding tasks. Therefore, we compare the performance when replacing CLIP features with DINOv2 features for node representation.

During registration, each 3DGS scene is rendered from multiple viewpoints, and features are extracted from the rendered images corresponding to each node. Graph association is then performed using both image feature similarity and spatial similarity to identify candidate correspondences between the two scene graphs. Subsequently, node matching is further refined based on the feature similarity. In this process, a key factor is how consistently the same object observed from different viewpoints is represented in the feature space.

As shown in Tab. 1, CLIP features produce more stable node matching results than DINOv2 features on the ScanNet-GSReg dataset [2], despite the latter being conceptually stronger image representations. This behavior can be attributed to differences in the representation characteristics learned by the two models. CLIP is trained on large-scale image-text pairs and tends to learn object-level semantic representations, which allows it to maintain relatively high feature similarity for the same object even when observed from different viewpoints. In contrast, DINOv2 is trained with self-supervised objectives and is more sensitive to fine-grained visual patterns and texture variations. As a result, appearance changes caused by viewpoint differences can lead to larger variations in the feature space, which may reduce feature similarity even for the same object.

We also observe that the cosine similarity between nodes corresponding to the same object across different viewpoints tends to be lower when using DINOv2 features. Our scene graph determines whether two nodes correspond to the same

**Table 2:** Runtime analysis on ScanNet-GSReg. Graph-GSReg performs online pairwise registration within 5 seconds after one-time offline graph construction.

| **Stage** | **Type** | **Runtime** |
|---|---|---|
| GaussReg (Training) | Offline | > 10 h |
| PhotoReg (Registration) | Online | > 5 min |
| 3DGS Rendering | | 0.007 s/view |
| Mask/feature extraction | Offline | 1.573 s/view |
| Graph association | | 0.251 s/view |
| *Ours Offline total* | | < **2 min** |
| Graph Registration | Online | 3.658 s |
| *Ours Online total* | | < **5 sec** |

object based on both feature similarity and spatial similarity, and performs association when both conditions are satisfied. Such lower similarity therefore makes it difficult to construct stable graph associations.

As a result, this issue propagates to the subsequent node matching stage, making it more difficult to recover correct node correspondences. Furthermore, the graph structure required for reliable registration often fails to be properly established, which ultimately leads to a lower overall registration success rate.

These characteristics are particularly important for cross-view object matching in graph-based registration. Consequently, CLIP features provide more reliable graph association and node matching, leading to more accurate graph-based scene registration.

## 5 Runtime Analysis of Graph Generation & Registration

Tab. 2 reports a detailed runtime breakdown of Graph-GSReg on ScanNet-GSReg, separating one-time offline preprocessing from online pairwise registration. Unlike previous methods that require time-consuming training or iterative optimization, Graph-GSReg constructs the scene graph within a short offline preprocessing time and performs efficient pairwise registration at test time.

In the offline stage, Graph-GSReg first renders multi-view images from each 3DGS scene, extracts object masks and visual features, and then associates them to construct an object-level scene graph. The overall offline preprocessing takes less than two minutes per scene. Among these steps, the runtime is mainly dominated by mask generation and feature extraction, which can be further reduced by adopting more efficient segmentation and feature extraction backbones.

After the scene graph is constructed, the online registration stage only requires graph matching and transformation estimation. As a result, Graph-GSReg completes pairwise registration within five seconds, demonstrating its efficiency for test-time 3DGS scene registration.

**Table 3:** Failure mode analysis under graph corruption on ScanNet-GSReg. We evaluate two practical corruption settings: removing object nodes to simulate missing masks and adding false candidate correspondences to simulate incorrect node matching.

| **Setting** | **RRE (°) ↓** | **RTE ↓** |
|---|---|---|
| Drop 30% Object Nodes | 3.620 | 0.044 |
| Add 30% False Matches | 3.300 | 0.042 |
| No Corruption | **3.247** | **0.039** |

## 6 Analysis of Graph Corruption and Failure Modes

To further analyze the robustness and limitations of Graph-GSReg, we evaluate the proposed method under two practical graph corruption settings. Since our method relies on object-level scene graphs, registration performance can be affected by imperfect graph construction, such as missing object nodes caused by incomplete masks or incorrect candidate correspondences caused by ambiguous object matching. We therefore simulate these two cases by deliberately corrupting the constructed scene graphs.

Specifically, in the first setting, we randomly remove 30% of object nodes from one of the two submaps. This setting simulates the case where object masks are missing or incomplete during scene graph construction. In the second setting, we inject 30% false matches into the initial candidate correspondences. This setting simulates incorrect node matching caused by visually similar or semantically ambiguous objects.

As shown in Tab. 3, both corruption settings lead to only minor degradation compared with the uncorrupted case. Even when 30% of object nodes are removed, Graph-GSReg maintains accurate registration with an RRE of 3.620° and an RTE of 0.044. Similarly, when 30% false matches are added, the method achieves an RRE of 3.300° and an RTE of 0.042. These results indicate that the proposed graph-based registration is robust to moderate graph construction errors.

This robustness mainly comes from the geometric consistency verification in TRIMs and the subsequent maximum clique selection. False or inconsistent correspondences are unlikely to satisfy pairwise distance consistency with many other correspondences, and are therefore suppressed during compatibility graph construction. As a result, the final clique is likely to contain geometrically reliable correspondences, allowing stable pose estimation even when the initial graph contains noisy matches.

Nevertheless, Graph-GSReg can still fail in more severe cases. Typical failure cases occur when the scene graph contains too few reliable object nodes, when most visible objects are repetitive or weakly distinctive, or when the overlap between two submaps is extremely limited. In such cases, the compatibility graph may not contain enough correct correspondences to form a reliable clique, which can lead to inaccurate transformation estimation. These observations suggest

**Table 4:** Fine registration results and overlap ratios for 69 sequence pairs in the uHumans2 [7] dataset.

| | Office | Apartment | Subway | Neighborhood | Mean |
|---|---|---|---|---|---|
| Overlap Ratio (%) | 42.34 | 38.70 | 37.08 | 20.07 | **34.55** |
| RRE (°) ↓ | 0.200 | 0.300 | 2.090 | 0.254 | **0.711** |
| RTE ↓ | 0.001 | 0.003 | 0.019 | 0.002 | **0.006** |
| ATE (m) ↓ | 0.021 | 0.019 | 0.557 | 0.173 | **0.192** |
| Time (s) ↓ | 3.485 | 2.466 | 3.810 | 5.367 | **3.782** |

that improving object-level graph construction under sparse, repetitive, or low-overlap conditions remains an important direction for future work.

## 7 3DGS Registration Dataset from uHumans2

ScanNet-GSReg [2] is one of the few publicly available datasets specifically designed for evaluating 3DGS registration. In addition, to evaluate registration performance in large-scale synthetic environments, we further construct a 3DGS registration test set using four sequences (office, apartment, subway, and neighborhood) from the uHumans2 [7] dataset.

Specifically, we extract RGB images, depth images, and IMU measurements at 5 FPS from the uHumans2 dataset, which is provided in rosbag format [5]. The IMU measurements are used to obtain camera poses for 3DGS reconstruction. From the extracted frames, we construct scenes consisting of 150 frames each, while ensuring that consecutive scenes share 50 overlapping frames, resulting in a total of 69 overlapped scene pairs.

As shown in Tab. 4, we report the average overlap ratio between scene pairs. ScanNet-GSReg, one of the few existing benchmarks for 3DGS registration, has an average overlap ratio of 56.23%, **whereas our dataset has a significantly lower average overlap ratio of 34.55%**, representing a more challenging setting with more limited overlap between scenes. These results demonstrate that the proposed method remains effective even under such conditions. The overlap ratio is measured using the original points.

## 8 Implementation Details

### 8.1 Details of 3D Scene Graph Construction

The 3D scene graph is constructed by first rendering each scene from multiple viewpoints, extracting object-level information from each frame, and then aggregating it into scene-level nodes. Specifically, SAM is applied to each rendered frame to obtain object masks. Only masks with confidence scores above a predefined threshold are retained, and CLIP features are extracted from the bounding box region of each mask to represent its semantic information. These mask-level

features are then used to associate objects observed across multiple frames within the same scene, which ultimately forms the nodes of the final scene graph. The detailed settings for graph association and the registration hyperparameters are summarized in Tab. 5.

Across all 82 scenes in ScanNet-GSReg, the average runtime of the scene graph construction process is approximately 2.076 seconds per frame for CLIP feature extraction, 1.089 seconds per frame for SAM-based mask extraction, and 0.007 seconds per frame for image rendering. In addition, graph association requires approximately 12.227 seconds per scene on average over the full benchmark. Most of this computational cost arises from object-level mask extraction and feature computation. However, since this stage is performed as an offline preprocessing step rather than during online inference, it remains practical within the overall pipeline. These runtimes may vary depending on factors such as rendering resolution, hardware configuration, and implementation details, and therefore should be interpreted as indicative rather than absolute measurements.

Furthermore, as described in Algorithm 1, scene graph construction does not require densely processing all frames, and can instead be performed reliably using temporally sampled sparse frames. This suggests substantial room for further acceleration through parallelization and lightweight implementations, making the overall process sufficiently efficient in offline settings.

**Table 5:** Hyperparameters for scene graph construction and registration.

| **Parameter** | **ScanNet-GSReg [2]** | **uHumans2 [7]** |
|---|---|---|
| Mask Confidence | 0.96 | 0.98 |
| Histogram Distance Threshold (m) | 1.0 | 1.0 |
| $\tau_{\text{asso}}$ | 1.7 | 1.8 |
| $\tau_{\text{TRIMs}}$ | 0.1 | 0.05 |
| $\tau'_{\text{TRIMs}}$ | 0.02 | 0.02 |
| Frame sampling interval ($s$) | 2 | 10 |
| Node Matching Threshold | 0.95 | 0.95 |
| Max ICP Iterations | 500 | 500 |
| ICP Distance Threshold (m) | 0.1 | 0.1 |

### 8.2 Details of TRIMs

To further analyze the effect of the TRIMs threshold $\tau_{\text{TRIMs}}$, we conduct an ablation study by varying the ratio from 0.1 to 0.9. Fig. 5 reports the resulting registration accuracy and runtime.

As the threshold increases, the distance consistency constraint becomes progressively relaxed, allowing more candidate correspondences to remain in the compatibility graph. While this increases the number of graph nodes, it also introduces more geometrically inconsistent matches, which degrades the registration accuracy. The enlarged compatibility graph further increases the computational cost of the maximum clique search.

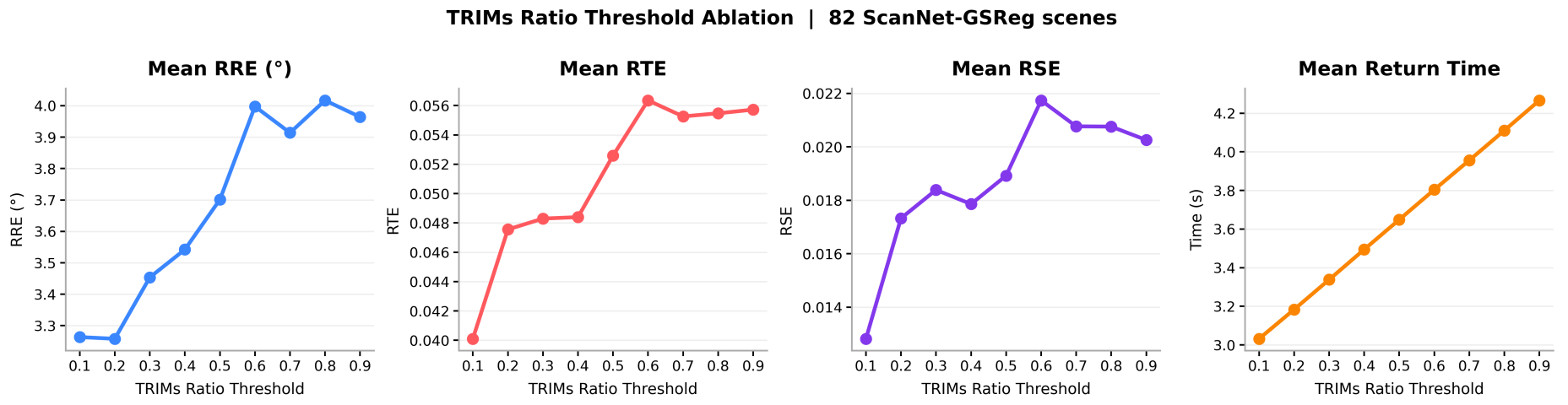


**Fig. 5: Ablation on the TRIMs threshold $\tau_{\text{TRIMs}}$.** We vary the threshold from 0.1 to 0.9 and evaluate the registration accuracy (RRE, RTE, RSE) and runtime. Larger thresholds relax the constraint, leading to lower accuracy and higher runtime.

Larger thresholds therefore lead to higher runtime and lower accuracy. Based on this observation, we set $\tau_{\text{TRIMs}} = 0.1$ in all experiments, as it provides the best trade-off between robustness and efficiency.

To prevent excessive computational overhead, we introduce an additional safeguard in Algorithm 1. When the compatibility graph becomes excessively large, the subsequent SVD verification step may incur substantial computational cost. In such cases, we intentionally reduce the threshold to $\tau'_{\text{TRIMs}}$ and apply TRIMs once more to regenerate the compatibility edges. This simple mechanism effectively limits the number of remaining correspondences, reducing the size of the compatibility graph and preventing the computational cost of subsequent steps from increasing excessively.

**Table 6:** Evaluation on 10 outdoor 3DGS scenes from Tanks and Temples [3]. Scenes with high symmetry or low texture are highlighted in bold. Graph-GSReg achieves accurate and efficient registration using only the graph-based coarse registration step.

| Scene | Method | RRE (°) ↓ | RTE ↓ | ATE (m) ↓ | Time (s) ↓ |
|---|---|---|---|---|---|
| Family, Horse, Lighthouse, M60, Panther, Playground, Train, **Francis**, **Museum**, **Temple** | TEASER++ & ICP | 27.177 | 0.268 | 0.633 | 8.199 |
| | PhotoReg (Coarse) | 32.940 | 0.701 | 1.531 | 30.703 |
| | PhotoReg (w/ Fine) | 31.304 | 0.693 | 1.272 | 337.738 |
| | Ours (Coarse) | **1.904** | **0.048** | **0.166** | **2.122** |

## 9 Evaluation in Challenging Outdoor Environments

To further evaluate the robustness of Graph-GSReg in challenging outdoor environments, we conduct an additional experiment on 10 outdoor 3DGS scenes from the Tanks and Temples [3] dataset. This benchmark includes scenes with large-scale structures, repetitive geometry, high symmetry, and weakly textured regions, which make appearance-based and photometric registration particularly challenging. Among the evaluated scenes, **Francis**, **Museum**, and **Temple** contain high symmetry or low-texture regions, and are therefore highlighted in Tab. 6.

As shown in Tab. 6, Graph-GSReg achieves substantially lower registration errors than the baseline methods while requiring only a few seconds. Notably, our method uses only the graph-based coarse registration step in this experiment, without additional fine-level photometric optimization. These results demonstrate that the proposed object-level graph representation provides robust structural cues even under challenging outdoor conditions where visual appearance alone can be ambiguous.

**Algorithm 1** Overall Pipeline of Graph-GSReg.

**Input:** 3DGS scenes $G^A, G^B$, Camera Poses $C^A, C^B$
**Output:** Optimized Merged 3DGS $G^{\text{merged}}_{\text{opt}}$, Transformation $\mathbf{T}^A_B$

**Phase 1: 3D Scene Graph Generation & Node Enrichment**
$C^A_{\text{sampled}} \leftarrow \{c \in C^A \mid \text{index}(c) \pmod{s} = 0\}$
$C^B_{\text{sampled}} \leftarrow \{c \in C^B \mid \text{index}(c) \pmod{s} = 0\}$
$\mathcal{G}^A, \mathcal{G}^B \leftarrow \text{3D_Scene_Graph_Generation}(G^A, G^B, C^A_{\text{sampled}}, C^B_{\text{sampled}})$
**for** each node $v \in \mathcal{V}^A \cup \mathcal{V}^B$ **do**
    $f(v) \leftarrow [f_{\text{CLIP}}(v), f_{\text{hist}}(v)]$ {Augment with 3-hop neighborhood histogram}
**end for**

**Phase 2: Compatibility Graph Construction & Registration**
$\mathcal{V}_{\text{comp}} \leftarrow \left\{(q,p) \mid \frac{f(q)\cdot f(p)}{\|f(q)\|\|f(p)\|} \geq \tau_{\text{node}},\ q \in \mathcal{V}^A, p \in \mathcal{V}^B\right\}$
$\mathcal{E}_{\text{comp}} \leftarrow \emptyset$
**for** each pair $m_1 = (q_1, p_1), m_2 = (q_2, p_2) \in \mathcal{V}_{\text{comp}}$ **do**
    $r \leftarrow \frac{\|q_1 - q_2\|}{\|p_1 - p_2\|}$
    **if** $1 - \tau_{\text{TRIMs}} < r < 1 + \tau_{\text{TRIMs}}$ **then**
        $\mathcal{E}_{\text{comp}} \leftarrow \mathcal{E}_{\text{comp}} \cup \{(m_1, m_2)\}$
    **end if**
**end for**
**if** $|\mathcal{E}_{\text{comp}}| > 1000$ **then**
    $\mathcal{E}_{\text{comp}} \leftarrow \emptyset$ {Apply strict TRIMs filtering}
    **for** each pair $m_1 = (q_1, p_1), m_2 = (q_2, p_2) \in \mathcal{V}_{\text{comp}}$ **do**
        $r \leftarrow \frac{\|q_1 - q_2\|}{\|p_1 - p_2\|}$
        **if** $1 - \tau'_{\text{TRIMs}} < r < 1 + \tau'_{\text{TRIMs}}$ **then**
            $\mathcal{E}_{\text{comp}} \leftarrow \mathcal{E}_{\text{comp}} \cup \{(m_1, m_2)\}$
        **end if**
    **end for**
**end if**
$\mathcal{G}_{\text{comp}} \leftarrow (\mathcal{V}_{\text{comp}}, \mathcal{E}_{\text{comp}})$
$\mathcal{C}_{\max} \leftarrow \text{Maximum_Clique}(\mathcal{G}_{\text{comp}})$
$\mathbf{T}_{\text{init}} \leftarrow \text{SVD}(\mathcal{C}_{\max})$
$P^A_{\text{sampled}}, P^B_{\text{sampled}} \leftarrow \text{Downsample_Points}(G^A, G^B)$
$\mathbf{T}^A_B \leftarrow \text{ICP}(P^A_{\text{sampled}}, P^B_{\text{sampled}}, \mathbf{T}_{\text{init}})$

**Phase 3: Merging & Test-Time Optimization**
$G^{B'} \leftarrow \text{Transform}(G^B, \mathbf{T}^A_B)$
$G^{\text{merged}} \leftarrow \text{Voxel_Downsample}(G^A \cup G^{B'}, \text{voxel_size} = 0.01)$
$G^{\text{merged}}_{\text{opt}} \leftarrow \text{Test_Time_Optimization}(G^A, G^{B'}, G^{\text{merged}})$
**return** $G^{\text{merged}}_{\text{opt}}, \mathbf{T}^A_B$

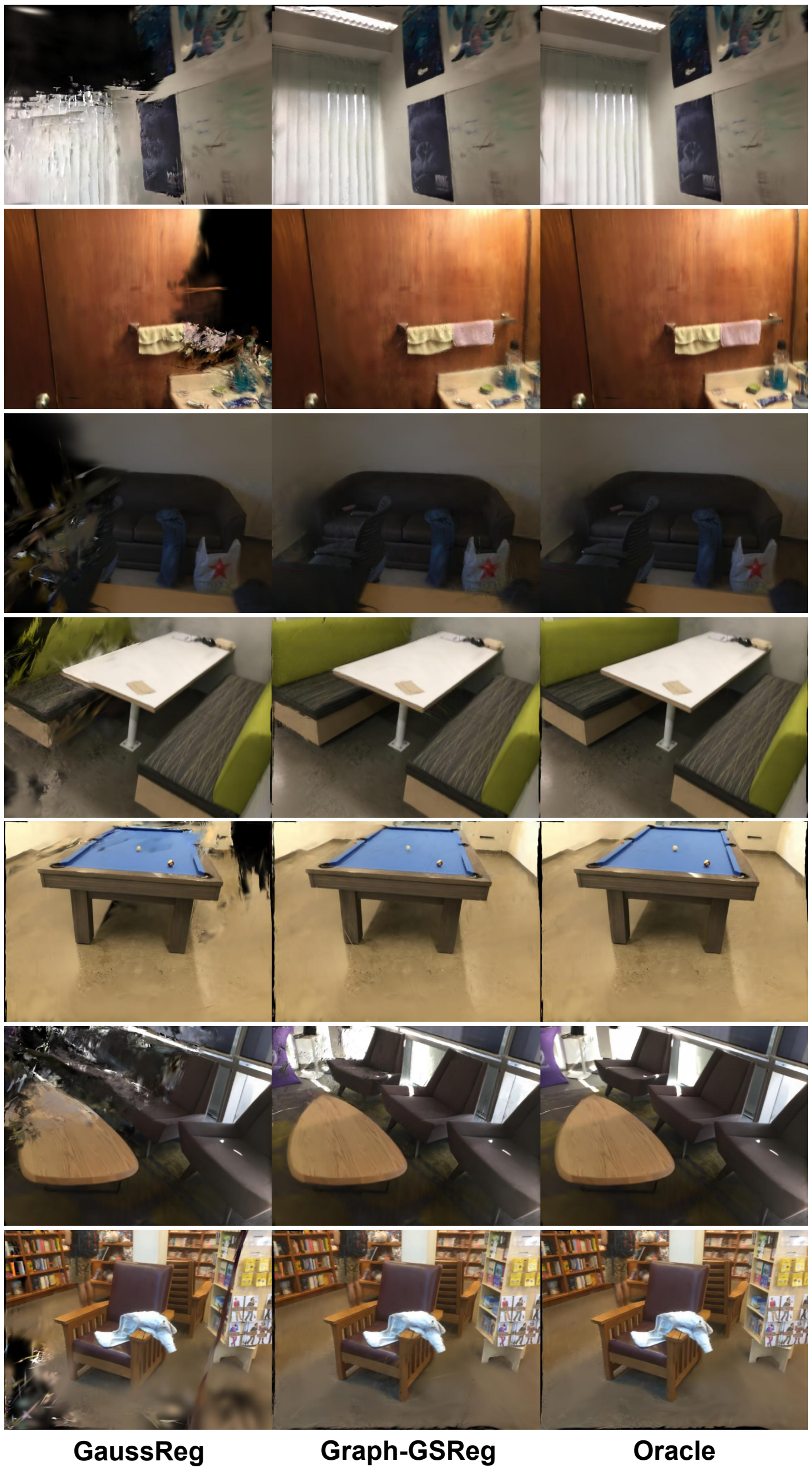


**Fig. 6: Qualitative results on ScanNet-GSReg [2].** To isolate the effect of the merging strategy, all methods use the same fixed registration result produced by our method. Since GaussReg [2] employs a distance-based heuristic for merging, it produces severe hollow artifacts. In contrast, our proposed method refines the merged scene through TTO, producing results close to the Oracle within a very short time while effectively avoiding hollow artifacts.

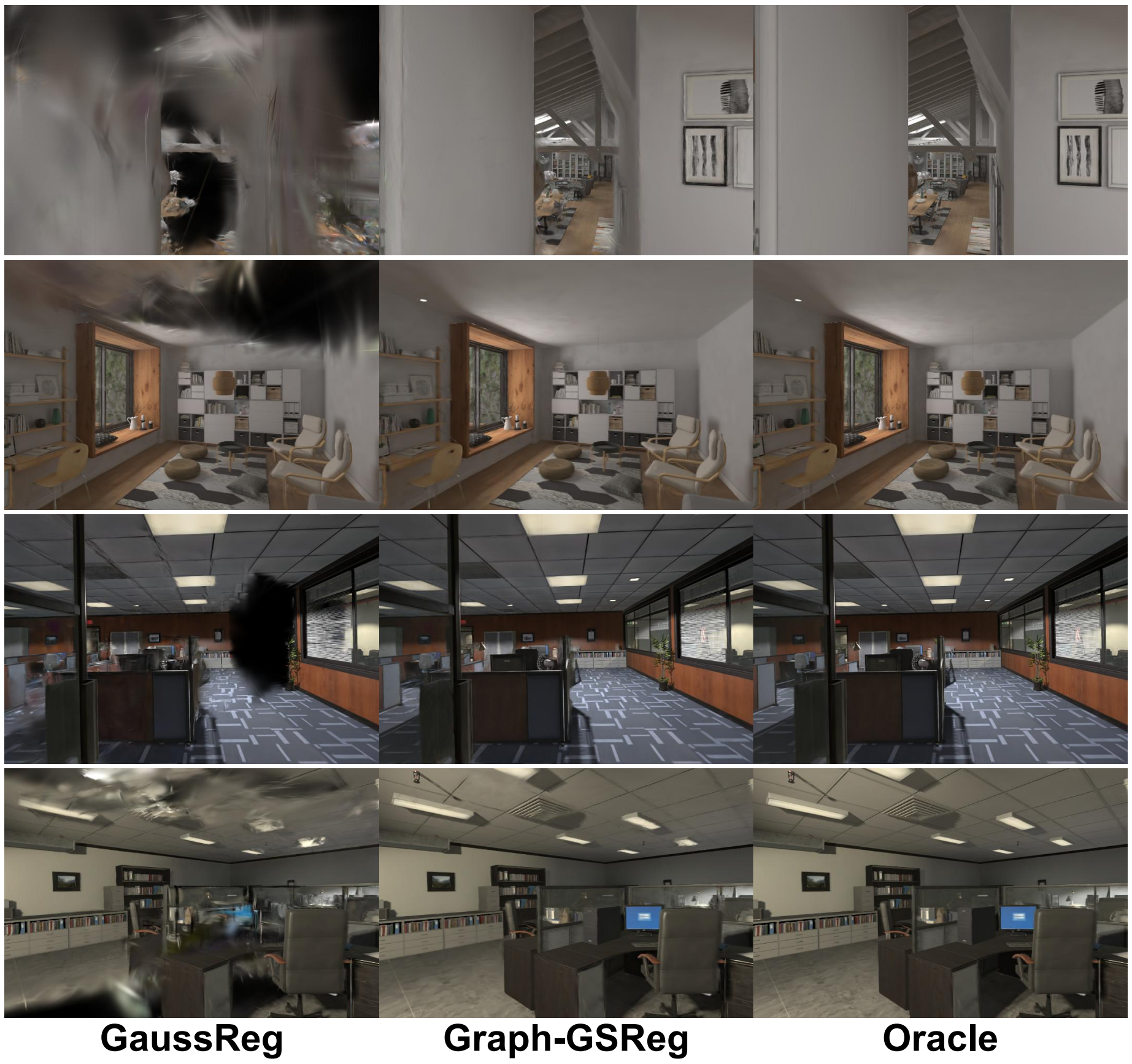


**Fig. 7:** Qualitative results on the Apartment and Office sequences from our uHumans2-based dataset.

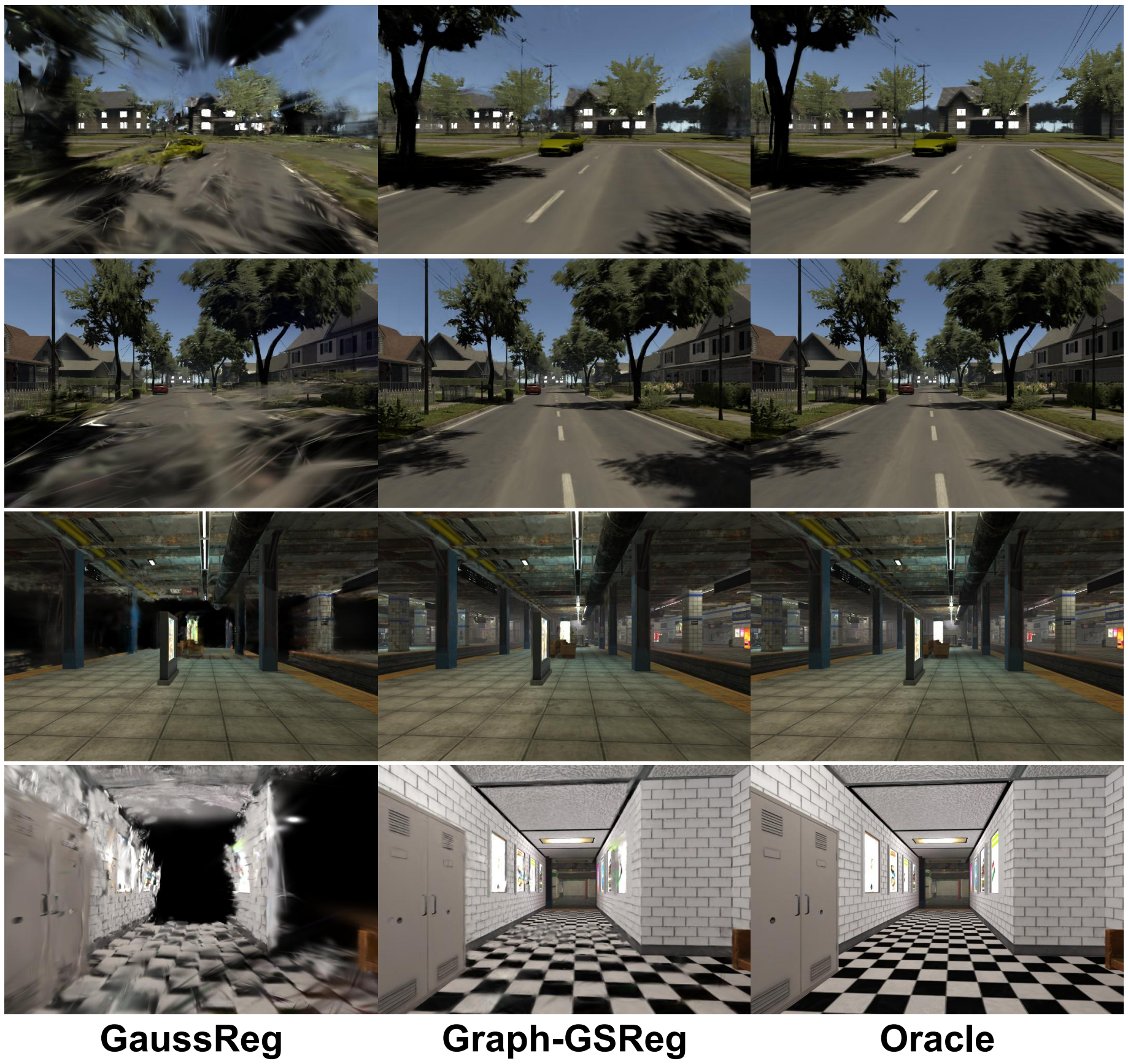


**Fig. 8:** Qualitative results on the Neighborhood and Subway sequences from our uHumans2-based dataset.



\end{document}